\RequirePackage{fix-cm}
\documentclass[twocolumn]{svjour3}          % twocolumn
\smartqed  % flush right qed marks, e.g. at end of proof
\usepackage{mathptmx}      % use Times fonts if available on your TeX system
\usepackage{yaacro}

\usepackage{float}

\usepackage{subcaption}
\usepackage{units}

\usepackage{booktabs}

\usepackage{longtable}

\usepackage{ifpdf}

\usepackage{xcolor}
\usepackage{graphics}
\usepackage[pdftex]{graphicx}
\DeclareGraphicsExtensions{.pdf,.PDF,.png,.PNG,.jpg,.JPG}
\usepackage{hyperref}
\hypersetup{
  colorlinks,%
  breaklinks=true,
  citecolor=red,%
  filecolor=red,%
  linkcolor=red,%
  urlcolor=red
}
\usepackage[
backend=biber,
style=authoryear,
maxnames=1,
maxbibnames=5,
maxcitenames=1,
dashed=false,
uniquelist=false,
citestyle=authoryear 
]{biblatex}
\bibliography{references}   % name your BibTeX data base

\newcommand*\rot{\rotatebox{90}}

\usepackage{amsmath}
\usepackage{amssymb}% http://ctan.org/pkg/amssymb
\usepackage{pifont}% http://ctan.org/pkg/pifont

\usepackage{siunitx,booktabs}

\makeatletter
\newcommand\argmin{\mathop{\operator@font arg\,min}}
\makeatother

\begin{document}
\sloppy

\title{Beyond Single Stage Encoder-Decoder Networks: Deep Decoders for Semantic Image Segmentation %\thanks{Grants or other notes
%about the article that should go on the front page should beencoderencoder
%placed here. General acknowledgments should be placed at the end of the article.}
}
%\subtitle{Do you have a subtitle?\\ If so, write it here}

%\titlerunning{Short form of title}        % if too long for running head

\author{Gabriel L. Oliveira    \and
        Senthil Yogamani \and
        Wolfram Burgard \and
        Thomas Brox %etc.
}

%\authorrunning{Short form of author list} % if too long for running head

\institute{Gabriel L. Oliveira \and Wolfram Burgard \and Thomas Brox \at University of Freiburg \\
              \email{\{oliveira,burgard,brox\}@cs.uni-freiburg.de} \\
           Senthil Yogamani \at
              Valeo Inc. \\
              \email{senthil.yogamani@valeo.com} \\   
}

\date{}
% The correct dates will be entered by the editor

\maketitle

\begin{abstract}
Single encoder-decoder methodologies for semantic segmentation are reaching their peak in terms of segmentation quality and efficiency per number of layers. To address these limitations, we propose a new architecture based on a decoder which uses a set of shallow networks for capturing more information content. The new decoder has a new topology of skip connections, namely backward and stacked residual connections. In order to further improve the architecture we introduce a weight function which aims to re-balance classes to increase the attention of the networks to under-represented objects. We carried out an extensive set of experiments that yielded state-of-the-art results for the CamVid, Gatech and Freiburg Forest datasets. Moreover, to further prove the effectiveness of our decoder, we conducted a set of experiments studying the impact of our decoder to state-of-the-art segmentation techniques. Additionally, we present a set of experiments augmenting semantic segmentation with optical flow information, showing that motion clues can boost pure image based semantic segmentation approaches.  

%Additionally, we introduce a novel network compression approach for multi-path convolutional blocks.   

\keywords{Deep Learning \and Semantic Segmentation \and Deep Decoder \and DPDB-Block \and Dynamic Weight Function \and Efficient Segmentation}
% \PACS{PACS code1 \and PACS code2 \and more}
% \subclass{MSC code1 \and MSC code2 \and more}
\end{abstract}

\section{Introduction}
\label{intro}

Deep learning approaches have become the standard for multiple perception tasks, 
like classification \parencite{VGG,Resnet,DenseNet}, object detection \parencite{FastRCNN}, optical flow \parencite{flownet2}, and 
semantic segmentation \parencite{long_shelhamer_fcn,unet,Deeplab,LRN17}. For dense prediction tasks, architectures that are build on Fully Convolutional Networks (FCNs) \parencite{long_shelhamer_fcn} have become the standard approach.  
%FCNs overcome limitations of previous approaches by incorporating context and using full resolution images as input, in contrast to sliding window semantic segmentation methods. 
These networks extend classification architectures, which consist solely of a convolutional encoder and some fully connected layers, to dense prediction by replacing the fully connected layers by a convolutional decoder that recovers the resolution lost by the down-sampling operation in the encoder. While largely used and efficient in their early adoption, single encoder-decoder networks are reaching a saturation in terms of segmentation quality and efficiency per number of layers. The main bottleneck of single stage decoders is that they cannot feed-back encoder layers with context information (decoder layers), such connection can make the architecture to extract more informative features. For example, when labeling a person image region, once the feature learning areas are aware that the region contains a person, the network may focus on the person-like visual patterns. 

In this paper, we aim to extend the decoder concept by proposing a new topology called Deep Decoder (DD). Deep decoders are decoder modules that not only upsample the features to a desired resolution, but also incorporate a feature learning capability to decoders by stacking multiple shallow decoder-encoder modules and connecting them in a way the shallow decoders are aware of the context information. 
We introduce new skip connection topologies and show that these improve the information flow, thus leading to better segmentation outputs while being computationally efficient. %We further approach common difficulties in deep encoder-decoder networks to propagate gradients by proposing new skip-connections and using deep supervision. 
%The new types of skip-connections incorporate contextual information to encoders and enhance information flow between decoders. 
Deep supervision is beneficial to train deeper architectures \parencite{deepsupervision}. In our setting, it is composed of a set of outputs in each decoder. With this multi-loss approach, we update the network in a hierarchical way, which improves the gradient propagation.  

%The decoder concept is combined with a new architecture block called Dual Path Dense Block (DPDB). This technique incorporates feature re-usage and new feature exploration capabilities into a single parameter efficient block. We presented DPDB in a preliminary conference version of this paper \parencite{oliveira2018icra}. A modified version of the DPDB block will be the main structure for the proposed encoder modules in the present paper.

The decoder (Deep Decoder) is the major new contribution of this work. It is further combined with the architecture block called Dual Path Dense Block (DPDB). DPDB blocks are designed to incorporate feature re-usage and new feature exploration capabilities into a single parameter efficient block. We presented the DPDB block in a preliminary conference version \parencite{oliveira2018icra}. A modified version, which is more focused on performance, is the main building block of the proposed encoder modules in this work.      

We also introduce a class balance weight function, which improves the network's attention to under-represented classes. The experimental evaluation shows that all the proposed measures lead to an approach that achieves state-of-the-art results on public datasets relevant for robotics, i.e, the CamVid \parencite{camvid}, Freiburg  Forest \parencite{Valada2017} and Gatech \parencite{Gatech} semantic segmentation datasets. % and Cityscapes \parencite{cityscapes} 

Network efficiency is a crucial aspect to multiple tasks, especially for robotics applications, due to computational limits on embedded hardware. Thus, we further improve on the efficiency of the approach by compressing the proposed architecture. We performed a set of experiments providing runtime values of our approach on multiple GPUs and investigate the best encoder feature learning block and deep decoder topology to provide the best trade-off between speed and segmentation quality, Section \ref{performance}. 

The remainder of the paper is organized as follows. We first discuss related work in Section~\ref{sec:relatedworks}. In Section~\ref{label:methodology}, we present an overview of the employed encoder block, propose our new decoder with it correspondent architecture and finally introduce our dynamic weight function for class balancing. Experimental results are reported in section~\ref{sec:experiments}. Finally, we summarize our work in Section~\ref{sec:conclusion}.

\section{Related Work}
\label{sec:relatedworks}
We review the recent advances in semantic segmentation using deep neural networks, represented by Fully Convolutional Networks (FCNs). FCNs \parencite{long_shelhamer_fcn} are composed by a fully convolutional topology and bi-linear interpolation to perform dense prediction. Following the FCN structure, many works try to alleviate problems related to rough edges segmentation and object vanishing, through exploring context, resolution and boundary alignment. Additionally, we will review efficient network architectures.

Context aggregation and retention of spatial information was explored by dilated convolutions and reduction of down-sampling operations \parencite{Dilated,Deeplab,DeeplabV2,GCN}. All approaches adopt dilated convolutions to enlarge the receptive field and capture larger contextual information without losing resolution. These methods also reduce the number of downsampling operation, such as pooling, in order to be able to have higher resolution feature maps at the end of the encoder, helping to produce more crispy edges. Recently, Deeplab-V2 proposes Atrous Spatial Pyramid Pooling, which combines features at different fields of view given by a set of dilated convolutions, to include context to a Resnet based encoder. Methods like Zoom-out \parencite{zoom-out} and ParseNet \parencite{parsenet} were designed to incorporate context explicitly. Zoom-out proposes a hierarchical context features network, while ParseNet includes global pooling features to explicitly add context information. Lately, the Global Convolutional Network (GCN) \parencite{GCN} incorporates context using large kernels to provide larger receptive fields.

Another set of approaches focusing on recovery of the resolution lost by down-sampling operations are Label Refinement Networks (LRN) \parencite{LRN17}, Deconv-Net \parencite{DeconvNet}, FC-Dense \parencite{FCDense} and DPDB-Net \parencite{oliveira2018icra}. LRNs introduces a multi-resolution refinement approach which solves the problem in a coarse-to-fine fashion by first predicting a low resolution semantic mask, then progressively refining the predictions to get a more detailed result. Each refinement is associated to a resolution related loss to improve information propagation over the network. Deconv-Net introduces an unpooling operation and an 
hourglass-like network to learn the upsampling process, while FC-Dense replaces the linear convolution operations by densely connected blocks \parencite{DenseNet}. 

Boundary approaches try to refine the predictions near the object edges. These approaches use the post-processing techniques, such as  Adelaide \parencite{Adelaide} and bilateral solver \parencite{BarronPoole2016}. Adelaide makes use of a CRF built on fully-connected graph, which serves as a boundary refinement after the CNN. Alternative solutions to CRFs are proposed by \parencite{BarronPoole2016,jampani2016}. \parencite{jampani2016} proposes the bilateral filter to learn specific potentials within CNNs, providing $10\times$ speed up and comparable performance to CRFs.

A range of studies has focused on exploring efficient convolutional networks that can be trained end-to-end, like Fast-Net \parencite{FastNet}, E-Net \parencite{ENet} and SegNet \parencite{segnet}. Fast-Net focuses on pruning over-parametrized layers targeting on efficiency in terms of computational requirements. E-Net introduces a deep convolutional encoder-decoder model with a residual bottleneck structure to build an efficient network architecture.   

\begin{figure*}[ht!] 
	\centering
	\includegraphics[width=1.0\textwidth]{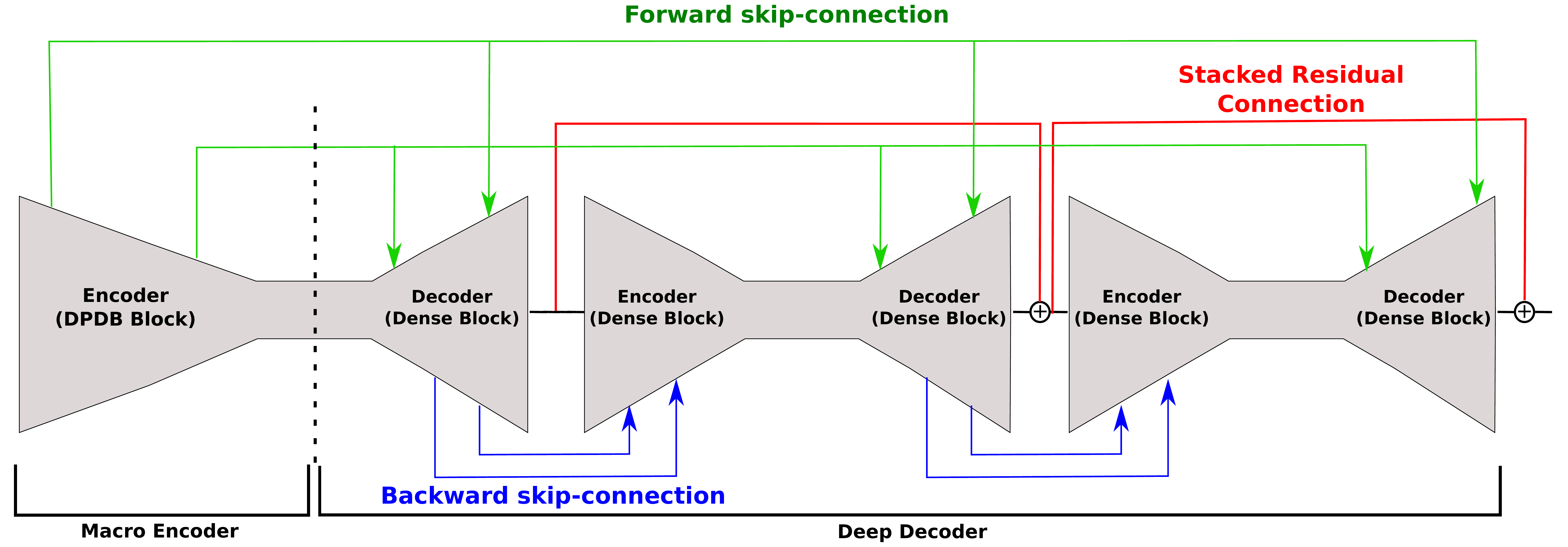}
	\caption{DD-Net Architecture. Our approach is composed by a macro encoder using DPDB blocks and our deep decoder, composed by sets of decoders-encoders. The main innovations reside in the backward and stacked residual connections. Backward connections aim to explicitly introduce context information to produce more informative features, while the stacked residual targets to enhance the information flow.} 
	\label{fig:DDnetFull}
\end{figure*}

Different from previous work, we will explore the potential of a new decoder which has a combination of shallow encoder-decoder networks to improve the description power of the proposed network. As highlighted in \parencite{wojna2017devil}, there is relatively lesser work done on segmentation decoders and it has become a bottleneck. We design an architecture to provide the most efficient computational requirement, given its highly deep topology. Additionally our dynamic weight function and deep supervision make our network easy to optimize and capable of producing more discriminative features. 

\section{The Proposed Approach}
\label{label:methodology}

We propose a decoder that includes multiple shallow networks and new skip connections between encoder-decoder, decoder-encoder, and decoder-decoder. These connections enable better information flow to deeper networks, and give encoders access to higher-level context information. In this section, we provide the details of our proposed method; the full architecture is presented in Figure \ref{fig:DDnetFull}. First, we briefly review the basic DPDB building block in the encoder. Second, we introduce the new decoder block that contains forward, backward, and stacked residual skip connections and deep supervision. Finally, we introduce a weight balancing function which  assigns weights dynamically and reinforce under-represented classes.  

\subsection{Dual Path Dense Block}
\label{subsec:dpdb}

In a preliminary conference paper \parencite{oliveira2018icra}, we introduced the dual path dense block (DPDB). It is an efficient subnetwork architecture that incorporates characteristics of feature re-usage and feature exploration to a single block. 
%An analysis of the main aspects of the based blocks to build the DPDB is presented, which is followed by an in-depth description of the module.    

\subsubsection{Analysis of ResNet and DenseNet}
\label{analysis}

DPDB is motivated by the strengths and weaknesses of the residual and densely connected topologies~\parencite{Resnet,DenseNet}, respectively. 

Let $x^l$ be the output of the $\textit{l-th}$ layer. Standard CNNs compute $x^l$ by applying a non-linear transformation $\phi_l$ to 
the output of the previous layer $x^{l-1}$. The equation $x^l = \phi_l(x^{l-1})$ defines $\phi_l$ as a set of operations, such as 
convolution followed by Exponential Linear Units (ELUs) \parencite{elus} and dropout. Residual networks introduced the so-called 
\textit{residual block} in order to ease the training of very deep architectures. The residual block sums the input and output layers: 

\begin{equation}
    x^l :=  \phi_l(x^{l-1}) + x^{l-1},
\end{equation}

making feature reuse possible and permitting gradients to flow directly to early layers. By sharing features across all steps, 
\textit{residual blocks} encourage feature re-usage and thus reduce feature redundancy. This makes it more difficult for residual 
networks to explore new features. For residual blocks, $N_l$ is usually defined as the repetition of $t$ blocks, usually two, 
composed by batch normalization, ReLU and convolution.

While \textit{residual blocks} repeat few blocks that are sequentially connected, DenseNets extend this idea with another type of architecture. Dense blocks recursively concatenates all previous feature outputs. The output $x_l$ of a DenseNet layer is defined as: 

\begin{equation}
    x^l :=  N_l([x^{l-1},x^{l-2}, x^{l-3}, ..., x^{0}]),
\end{equation}

%\textcolor{red}{where each layer is a composition of all previous ones through concatenation $[\cdots]$. Given a recurrent neural network (RNN) view to densely connected blocks, considering $h^t$ the hidden state of the $t$-th step of a RNN and $k$ the index of the current step. For each step, $f_t^k(.)$ is the feature extraction function which is responsible for taking a hidden state $h^t$ as input and outputting the extracted information. Additionally, a gathered information function $g^k(.)$ is denote as a function to transform  the current hidden state $h^k = g^k[f_t^k(h^t)]$. We can consider that for each micro-block step dense blocks are 
%able to explore new information from previous outputs since the functions $f_t^k(.)$ and $g^k(.)$ are not shared across steps. However, different $f_t^k(.)$ can potentially extract similar features multiple time, resulting on high redundancy.} 

where each layer is a composition of all previous ones through concatenation $[\cdots]$. The main characteristic of densely connected blocks is the ability to explore new information from previous outputs \parencite{DPN}. Hence, different features may extract the same information multiple times, leading to a high redundancy block.

The residual network's main limitation is its summation operation for fusing information. This operation may 
squash useful features from preceding layers. The squashing problem can be interpreted as follows: given two vectors of weights 
$\mathbf{w}_1 \triangleq [w_{11},...,w_{1n}]$ and $\mathbf{w}_2 \triangleq [w_{21},..., w_{2n}]$ and an element-wise aggregation 
function $f_{ag}(\mathbf{w}_{1}, \mathbf{w}_{2}) \triangleq \mathbf{w}_1 + \mathbf{w}_2$, thus, if $w_{1j} \gg w_{2j}$ then 
$f_{ag}(\mathbf{w}_1,\mathbf{w}_2) \sim \mathbf{w}_1 + \pmb{\epsilon}$ then the importance of the low magnitude weights vanishes. 
Additionally, its high number of parameters, makes very deep residual networks intractable. DenseNets on the contrary can provide 
a better efficiency in term of parameter usage considering the block operations. On the other hand, dense blocks have an excessive parameter growth, due to the 
characteristic of successive dense blocks always incorporate the full feature size of the input to compose the new output feature map, which limits the width of DenseNets.       

In the following section we will present the Dual-Path Dense-Block (DPDB) approach which combines the advantages of both architectures in a single block.

\subsubsection{DPDB Block}

Based on the previous analysis, we propose a new dense block called Dual-Path Dense-Block. Our block is different from the dual path network \parencite{DPN}, which also combines concepts from ResNet and DenseNet: we give similar weights to each of the sides and do not use a residual block as main block adding a thin densely connect path.

Given $x_{l,R}$ and $x_{l,D}$ as the outputs for the $\textit{l-th}$ layers of the residual path and dense path, we formulate the DPDB path block as: 
\begin{align}
& x_{l,R} := f_l^t \left ( x_{l,R}^t \right ) = x_{l,R}^{l-1} + \phi_l^t \left ( x_{l,R}^{l-1} \right ),  \label{residualpath}\\
& x_{l,D} := \sum_{t=0}^{l-1} N_l^t \left ( \left [ x_{l,D}^t \right ] \right ), \label{densepath}\\
& r_l := \left [x_{l,R}, x_{l,D} \right ], \label{fusion}\\
& h_l :=G_l(r_l), \label{final_transformation}
\end{align}

where $f_l^t$ and $N_l^t$ are the feature learning function. Equation \ref{residualpath} refers to the residual path that enables feature re-usage and Equation \ref{densepath} to the densely connected path that enables new feature exploration. Equation \ref{fusion} defines the path that fuses the outputs and feeds them to the final transformation function in 
Equation \ref{final_transformation}. The transformation function $G_l(\cdot)$ is responsible for making the next mapping or prediction. Path fusion is done by concatenation to avoid the feature squashing problem.

\begin{figure}[ht!] 
	\centering
	\includegraphics[width=.7\linewidth]{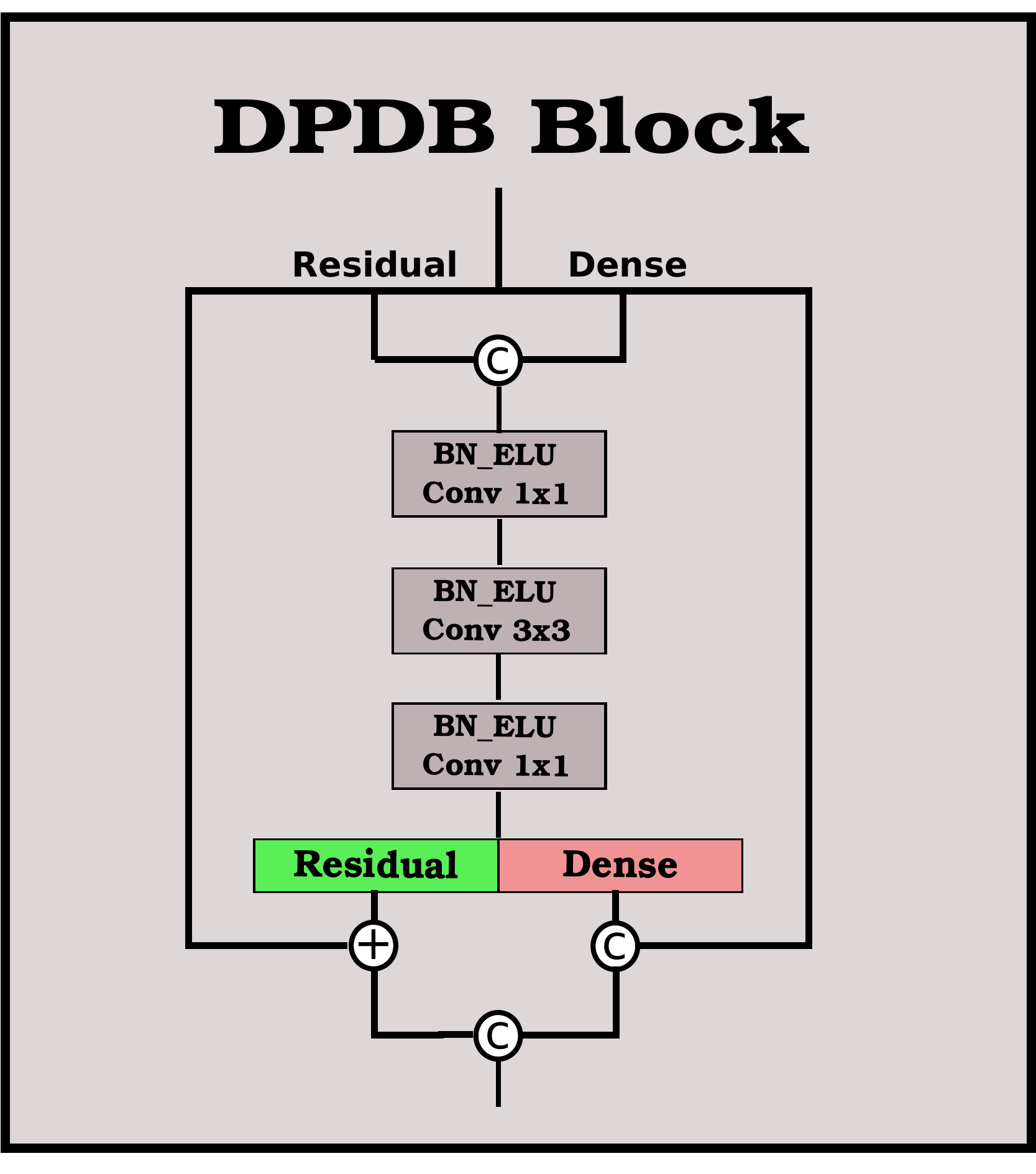}
	\caption[Dual-Path Dense-Block]{DPDB block. Dual-Path Dense-Block (DPDB) is a feature learning approach which incorporates feature re-usage (Residual Block) and new feature exploration (Dense Block) into a single learning scheme. BN stands for batch normalization and ELU for exponential linear unit.}
	\label{fig:dpdbblock}
\end{figure}

A full description of the proposed block is depicted in Figure \ref{fig:dpdbblock}. The block consists of two paths that pass through a bottleneck layer. The latter consists of batch normalization, ELU activation function and convolution, followed by a 3x3 kernel layer and finally to a output dimension specific layer. The output is then split into its corresponding path which will employ the specific aggregation functions. 

\subsection{Deep Decoder}
\label{subsec:decoder}

We propose a new decoder topology, which aims to recover high-resolution predictions through a set of multiple shallow encoder-decoder networks, as shown in Figure \ref{fig:DDnetFull}. Each set of networks are piled up from end-to-end, and forward, backward, and stacked residual connections are jointly employed. 
%The proposed connected structure of the network enables a more efficient backward gradient flow, and more detailed boundary restoration with better efficiency given the decoder depth. 
The structure of each set consists of a decoder-encoder which adopts Dense Blocks and three upsampling blocks followed by three downsampling ones. In order to further improve information flow and discriminability in the network we deeply supervise all the last units of the decoder block.

\subsubsection{Skip Connections}
\label{subsec:decoderblock}

%Decoder blocks are concatenated in our network to enhance the learning power of our approach. 
The architecture makes use of three different skip connections, namely forward, backward and stacked residual. Forward skip connections are responsible for associating features from the first encoder to all subsequent decoders, backward connections link two adjacent decoder-encoder units while the stacked residual connections work like macro residuals between decoders.

The first type of skip connection is the standard forward one. Forward skip connections connect parts of the encoder with its respective same resolution decoder counterpart, this is useful given the mid-level representations from the first encoder not pass through a series of downsampling operation that usually reduces its spatial visual information. Thus, such connection promotes the inclusion of less corrupted spatial information and consequently produces better detailed boundaries predictions. The skip-connection operation consists of feeding the features from the layer $n^{th}$ of first encoder into a convolutional layer to reduce the number of feature maps to $F^{n}_1$. The reduced features $F^{n}_1$ are then aggregated to the output of the upsampling operation through concatenation. The upsampling operation is composed by a transposed convolution followed by a Dense Block, further discussion about this choice is done in Section \ref{featurelearingdecoder}.            

The second type of connection is the backward skip connection, which is inspired by \parencite{backwardskip}. The authors proposed a connection from late convolutions to early convolution layers in order to enable the early convolution layers understand the context earlier and, thus, being able to adaptively extract more informative features. \parencite{backwardskip} proposed a master-slave network where the master network is responsible for producing the label prediction while the slave only provides the backward skip connections to the master's earlier layers. Their skip connection suffers from a major drawback, which is the doubling of computational costs. Our backward skip connection has similar effects, but with a much smaller computational burden. Instead of a master-slave architecture, we use a set of encoder-decoder networks, where each new encoder in the macro decoder block has backward skip-connections from its previous decoder. This promotes the flow of high-level semantic information to succeeding encoder layers, improves the encoder optimization, and, consequently, makes the context to be captured early on these parts of the network. We adopt three backward connections between every decoder-encoder of DD-Net, using element-wise summation as aggregation function. Three skip connections are related to the three blocks used in each encoder, which are consequently related to the $8x$ dimensionality reduction which occur at each of the modules.     

Along with forward and backward connections we install skip connections between all the high resolution outputs of each decoder. We call this Stacked Residual Connection (SRC). The main goal of SRCs is to act like a macro residual connection between each encoder-decoder set and additionally to produce a multi-stage segmentation mask prediction. From the second decoder on we consider the $n^{th}$ final decoder layer, the output $F_{i}^{n}$ are fused with features $F_{i}^{n-1}$ from the previous final decoder layer by element-wise summation to produce a new fused feature $\bar{F}_{i}^n = F_{i}^{n} \bigoplus F_{i}^{n-1}$ which then serve as input to the next encoder-decoder block. Such connection can enhances information flow and improve segmentation results.

\subsubsection{Deep Supervision}
\label{subsec:supervision}

%Deep architectures can potentially produce better results. Nevertheless, deeper networks face gradient propagation as one of its main problems. Our architecture given the large number of layers can experience additional optimization difficulties. Forward, backward and stacked residual connections can assist information flow but not completely mitigate such problem. In order to further reduce this problem, we add deep supervision \parencite{deepsupervision} to each encoder-decoder module. The proposed supervision acts like producing prediction on each final decoder layer and helps to optimize the learning process.        

Deeper architectures can potentially produce better results, yet with deeper networks comes the issue of gradient propagation. Skip-connections, like backward and stacked residual, can improve the information flow, but does not completely mitigate such problem. In order to further reduce this problem, we make use of deep supervision \parencite{deepsupervision}.

Deep supervision consists of adding auxiliary supervision branches after certain intermediate layers during training. One of the key aspects of deeply supervised training is where to add the supervision branches. We add auxiliary supervision to the end of each decoder block. The combined loss function for the whole network is then composed by $K$ auxiliary losses and a main loss, where $K$ is the number of decoder minus one.

%\subsection{Network Sparsification}
%\label{subsec:Sparsification}

%Separable convolutions are a special case of a sparsely connected convolution. Standard convolutions generate $F$ output 
%features, 
%by the application of a convolution filter over all $I$ input features. The operation $I \times F$ can be prohibitively expensive 
%if the number of input features is large. Separable convolutions divide the partition into $G$ mutually exclusive groups, each 
%reducing the cost by a factor of $G$ to $\frac{I \times F}{G}$. Such approach have proven gains for classification architectures, 
%as firstly adopted in the AlexNet architecure \parencite{Alexnet}, however for densely connected architectures and at decoder in fully %convolutional networks such technique is not fully explored.            

\subsection{Dynamic Weight Function}
\label{subsec:weightFunction}

Semantic segmentation tasks often come with an uneven distribution of classes in images. A possible solution to class balancing is provided by weight functions. Weighting functions are responsible for providing a new class distribution to the cross-entropy loss in order to make all classes equally important. One strong characteristic from most segmentation input images is the usually low number of instances of a single class per example. Thus, smaller objects will most probably be less represented per sample. This trait  motivated the introduction of a new weight function that reweights the class distribution to improve the network attention to these classes, which are often neglected by approaches which consider each class equal in importance.

Median frequency \parencite{Eigen15} have become the standard method to produce balanced weights for semantic segmentation. However effective for some segmentation problems, it requires to access the whole dataset prior to training and produces a static weight distribution to every sample. This limits its application domain and performance.

%One possible solution to class balancing is median frequency \parencite{Eigen15}. Median frequency needs to have 
%access to the entire dataset before training and produce a static weight distribution. The weights produce by such technique show better results than an unbalanced network, however the static weight computation and similar weights between different target objects and background greatly limits its application.

Thus, we propose a novel weight function, which assigns weights dynamically, does not need any pre-processing step, and reinforces classes that are under-represented by standard class balancing approaches. We aim to produce a set of weights that are a trade-off between dominant classes and less-represented ones. Our function gives an inverse weight using as base the class pixel frequency, i.e., smaller objects will have higher weights:

\begin{equation}
    DW_i = \frac{C_b + \sum_{i=1}^{N} C_i}{C_i}
    \label{weight}
\end{equation}

where $N$ is the number of classes, $C_b$ is the number of background pixels in the image and $C_i$ the number of pixels of the class $i$. 

The downside of this function is that frequent classes drop in performance. To restrict this effect, we limit the weight to always be greater than a constant $L$:

\begin{equation}
    DW_{bounded,i} = max(DW_i, L)
    \label{weight_LOWER_BOUND}
\end{equation}

where $DW_{bounded,i}$ is the lower-bounded weight for class $i$, thus we can guarantee no class weight is smaller than one since we further divide $DW_{bounded,i}$ by $L$. We set the background weight always to zero.

\section{Experiments}
\label{sec:experiments}

We evaluated the performance of our network on three common robotics datasets, CamVid
dataset \parencite{camvid}, the Freiburg Forest dataset \parencite{Valada2017} and the Gatech context dataset \parencite{Gatech}.
The implementation was based on the publicly available TensorFlow learning toolbox [1] and all experiments
were carried out on an NVIDIA Titan X GPU.

%http://lmb.informatik.uni-freiburg.de/resources/binaries/.

For the compared datasets, we report quantitative results and benchmark them with state-of-the-art baselines. We present an in-depth ablation study related to the impact of feature learning at the decoder, class balance approaches, depth of the proposed decoder and network pre-training.

Moreover, we conducted experiments studying the impact of deep decoders to state-of-the-art segmentation techniques, quantifying the gain of using our approach as decoder. As a final set of experiments augmenting semantic segmentation with optical flow information is presented. 

\begin{table*}[ht]
\centering
\caption{Results on CamVid dataset. Our approach obtained an average mIoU of $73.2$ percentage points, constituting the new state-of-the-art for the present dataset. Additionally, we only require one fifth of the number of parameters of the second best method.}
\label{camvidTable}
\begin{tabular}{l|c||c @{\hspace{0.8\tabcolsep}}|c @{\hspace{0.8\tabcolsep}}|c @{\hspace{0.8\tabcolsep}}|c@{\hspace{0.8\tabcolsep}}|c@{\hspace{0.8\tabcolsep}}|c@{\hspace{0.8\tabcolsep}}|c@{\hspace{0.8\tabcolsep}}|c@{\hspace{0.8\tabcolsep}}|c@{\hspace{0.8\tabcolsep}}|c@{\hspace{0.8\tabcolsep}}|c@{\hspace{0.8\tabcolsep}}||c@{\hspace{0.8\tabcolsep}}|c}
Method & \rot{Parameters (M)} &  \rot{Building}  & \rot{Tree} &  \rot{Sky} &  \rot{Car} &  \rot{Sign} &  \rot{Road} &  
\rot{Pedestrian} &  \rot{Fence} &  \rot{Pole} &  \rot{Sidewalk} &  \rot{Cyclist} &  \rot{Global Acc.} & \rot{mIoU} \\
\toprule
\toprule
S.parsing {\scriptsize\parencite{superparsing}} & $-$ & $70.4$ & $54.8$ & $83.5$ & $43.3$ & $25.4$ & $83.4$ & $11.6$ & $18.3$ & $5.2$ & $57.4$ & $8.9$ & $-$ & $42.0$\\
ALE  {\scriptsize\parencite{ALE}}   & $-$ & $73.4$ & $70.2$ & $91.1$ & $64.2$ & $24.4$ & $91.1$ & $29.1$ & $31.0$ & $13.6$ & $72.4$ & $28.6$ & $-$ & $53.6$\\
Liu {\scriptsize\parencite{Liu_2015_CVPR}}  & $-$ & $66.8$ & $66.6$ & $90.1$ & $62.9$ & $21.4$ & $85.8$ & $28.0$ & $17.8$ & $8.3$ & $63.5$ & $8.5$ & $-$ & $47.2$\\
SegNet {\scriptsize\parencite{segnet}}   & $29.5$ & $68.7$ & $52.0$ & $87.0$ & $58.5$ & $13.4$ & $86.2$ & $25.3$ & $17.9$ & $16.0$ & $60.5$ & $24.8$ & $62.5$ & $46.2$\\
DeconvNet {\scriptsize\parencite{DeconvNet}} & $252$ & $-$ & $-$ & $-$ & $-$ & $-$ & $-$ & $-$ & $-$ & $-$ & $-$ & $-$ & $85.9$ & $48.9$\\
FCN8s {\scriptsize\parencite{long_shelhamer_fcn}} & $134$ & $77.8$ & $71.0$ & $88.7$ & $76.1$ & $32.7$ & $91.2$ & $41.7$ & $24.4$ & $19.9$ & 
$72.7$ & $31.0$ & $88.0$ & $57.0$\\
STFCN {\scriptsize\parencite{STFCN}}  & $-$ & $73.5$ & $56.4$ & $90.7$ & $63.3$ & $17.9$ & $90.1$ & $31.4$ & $21.7$ & $18.2$ & $64.9$ & 
$29.3$ 
& $-$ & $50.6$\\
Reseg {\scriptsize\parencite{Reseg}}  & $-$ & $-$ & $-$ & $-$ & $-$ & $-$ & $-$ & $-$ & $-$ & $-$ & $-$ & $-$ & $88.7$ & $58.8$\\
LRN {\scriptsize\parencite{LRN17}}  & $-$ & $-$ & $-$ & $-$ & $-$ & $-$ & $-$ & $-$ & $-$ & $-$ & $-$ & $-$ & $-$ & $61.7$\\
Bayesian {\scriptsize SegNet \parencite{BayesianSegnet}}  & $29$ & $-$ & $-$ & $-$ & $-$ & $-$ & $-$ & $-$ & $-$ & $-$ & $-$ & $-$ & $86.9$ & 
$63.1$\\
DeepLab{\scriptsize-LFOV \parencite{Deeplab}}  & $37.3$ & $81.5$ & $74.6$ & $89.0$ & $82.2$ & $42.3$ & $92.2$ & $48.4$ & $27.2$ & $14.3$ & $75.4$ & 
$50.1$ & $-$ & $61.6$\\
Dilation {\scriptsize\parencite{Dilated}}  & $140$ & $82.6$ & $76.2$ & $89.0$ & $84.0$ & $46.9$ & $92.2$ & $56.2$ & 
$35.8$ & $23.4$ & $75.3$ & $55.5$ & $79.0$ & $65.3$\\
FCCN {\scriptsize\parencite{fccn}}  & $-$ & $79.7$ & $77.2$ & $85.7$ & $86.1$ & $45.3$ & $94.9$ & $45.8$ & $\bf{69.0}$ & $25.2$ & $86.2$ & $52.9$ & $-$ & $65.7$\\
Kundu {\scriptsize\parencite{KunduCVPR16}} & $-$ & $84.0$ & $77.2$ & $91.3$ & $85.6$ & $49.9$ & $92.5$ & $59.1$ & $37.6$ & $16.9$ & $76.0$ & $57.2$ & $-$ & $66.1$\\
FC-DenseNet103 {\scriptsize\parencite{FCDense}} & $-$ & $83.0$ & $77.3$ & $93.0$ & $77.3$ & $43.9$ & $94.5$ & $59.6$ & $37.1$ & $37.8$ & $82.2$ & $50.5$ & $91.5$ & $66.9$\\
DCNN {\scriptsize\parencite{Fu2017DenselyCD}} & $-$ & $-$ & $-$ & $-$ & $-$ & $-$ & $-$ & $-$ & $-$ & $-$ & $-$ & $-$ & $91.4$ & $68.4$\\
G-FRNet {\scriptsize\parencite{Islam_2018}} & $-$ & $83.7$ & $77.8$ & $92.5$ & $83.6$ & $44.7$ & $94.6$ & $58.4$ & $45.2$ & $34.7$ & $83.2$ & $58.1$ & $-$ & $68.8$\\
Playing for data {\scriptsize\parencite{Richter16}} & $-$ & $84.4$ & $77.5$ & $91.1$ & $84.9$ & $51.3$ & $94.5$ & $59.0$ & $44.9$ & $29.5$ & $82.0$ & $58.4$ & $-$ & $68.9$\\
SDN {\scriptsize\parencite{SDN}} & $161$ & $85.2$ & $77.5$ & $92.3$ & $\bf{90.2}$ & $\bf{53.9}$ & $96.0$ & $63.8$ & $39.8$ & $38.4$ & $85.3$ & $66.9$ & $92.7$ & $71.8$\\
\midrule
\midrule
%Dense blocks {\scriptsize\parencite{oliveira2018icra}}  & $36.2$ & $66.9$ & $62.4$ & $81.1$ & $45.5$ & $28.8$ & $83.2$ & $38.9$ & $26.8$ %& 
%$22.5$ & $62.0$ & $22.4$ & $81.0$ & $49.1$\\
%DPDB blocks {\scriptsize\parencite{oliveira2018icra}}  & $55.2$ & $72.4$ &  $62.9$ & $88.6$ & $61.9$ & $30.0$ & $88.8$ & $44.8$ & $26.1$ %& 
%23.6$ & $69.4$ & $33.1$ & $85.2$ & $54.7$\\

%Ours  & $46.4$ & $79.5$ & $74.7$ & $\bf{92.5}$ & $79.3$ & $41.3$ & $93.8$ & $54.0$ & $22.8$ & $\bf{32.7}$ & $78.8$ & $58.8$ & 
%$90.2$ & $64.4$\\

%Ours No Class Balance & $46.4$ & $81.0$ & $74.5$ & $92.4$ & $80.9$ & $42.4$ & $\bf{94.7}$ & $\bf{58.8}$ & $17.8$ & $31.5$ & 
%$\bf{80.6}$ & $54.6$ & $\bf{90.9}$ & $64.5$\\

%Ours & $31.6$ & $80.4$ & $74.0$ & $\bf{92.1}$ & $78.8$ & $41.3$ & $\bf{94.5}$ & $\bf{56.9}$ & $25.7$ & $\bf{34.1}$ & $\bf{80.7}$ 
%& $\bf{61.8}$ & $\bf{90.4}$ & $\bf{65.5}$\\

%$Ours & $31.6$ & $81.5$ & $75.1$ & $\bf{92.1}$ & $80.1$ & $43.1$ & $\bf{95.0}$ & $58.8$ & $28.3$ & $\bf{35.4}$ & $82.2$ & %$\bf{58.1}$ & $\bf{91.0}$ & $\bf{66.3}$\\

DD-Net & $31.6$  & $\bf{85.3}$ & $\bf{79.4}$ & $\bf{93.0}$ & $86.7$ & $51.4$ & $\bf{96.7}$ & $\bf{68.5}$ & $41.1$ & $\bf{44.6}$ & $\bf{88.0}$ & $\bf{70.4}$ & $\bf{93.1}$ & $\bf{73.2}$\\

\bottomrule
\end{tabular}
\end{table*}

\subsection{Architecture and training details}

The network was trained end-to-end using the Adam solver \parencite{AdamSolver} with an initial 
learning rate of $2 \times 10^{-4}$ which decay $10\times$ every $2\times10^{5}$ iterations. All models were trained on data 
augmented images with multi-window random crop and vertical flip. We also applied mean subtraction to images and weight class balancing and regularized our model with a weight decay of $10^{-4}$ and a dropout rate of $0.1$. The mean IoU was monitored every $100$ iterations.

\vspace{-0.3cm}

\subsection{CamVid dataset}

CamVid is a dataset of fully segmented videos for semantic segmentation of urban environments \parencite{camvid}. The dataset is constituted by $367$ frames for training, $101$ frames for validation and $233$ frames for testing. Each frame has $360\times480$ pixels, which are labeled with 11 semantic classes. We trained our network with augmented frames and fine-tuned using a model pre-trained on Cityscapes. The impact of these design choices is quantified in the ablation study in Section \ref{subsec:ablation}. 

The results are summarized in Table \ref{camvidTable}. As shown in the table, CamVid is an actively benchmarked dataset, the compared methods range from FCNs with dilation convolutions like \parencite{Dilated}, networks with Dense Blocks such as FCDenseNet \parencite{FCDense}, with additional training data \parencite{Richter16} to even deeper architectures like SDN \parencite{SDN}. From all the compared methods only SDN was capable of surpassing $70$ percentage points of intersection over union. Our approach not only outperforms SDN and defines the new state-of-the art on CamVid, but DD-Net also uses five times fewer parameters than SDN. 

\begin{figure*}
    \centering
    {
    \begin{subfigure}{.30\linewidth}
           \includegraphics[width=\linewidth]{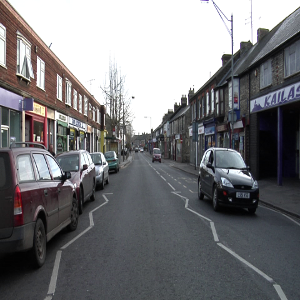}
    \end{subfigure}
    \begin{subfigure}{.30\linewidth}
       \includegraphics[width=\linewidth]{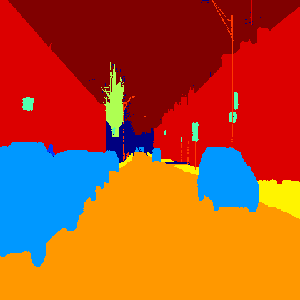}
    \end{subfigure}
    \begin{subfigure}{.30\linewidth}       
        \includegraphics[width=\linewidth]{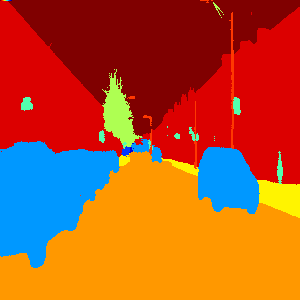}
    \end{subfigure}    
    } \\
    {
    \begin{subfigure}{.30\linewidth}
	\includegraphics[width=\linewidth]{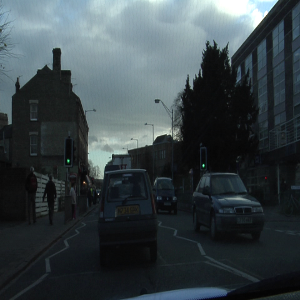}
	\end{subfigure}
    \begin{subfigure}{.30\linewidth}
  	\includegraphics[width=\linewidth]{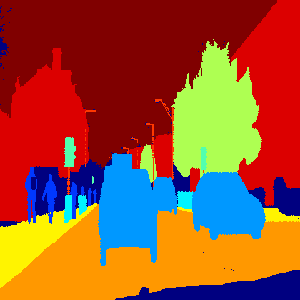}
    \end{subfigure}
    \begin{subfigure}{.30\linewidth}  	
  	\includegraphics[width=\linewidth]{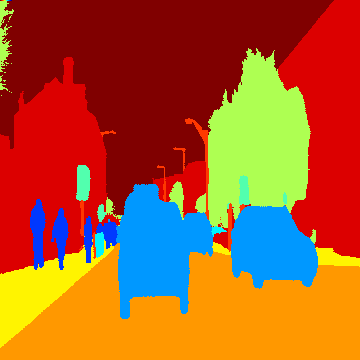}
  	\end{subfigure}  
    }
    \\
    {
    \begin{subfigure}{.30\linewidth}
	\includegraphics[width=\linewidth]{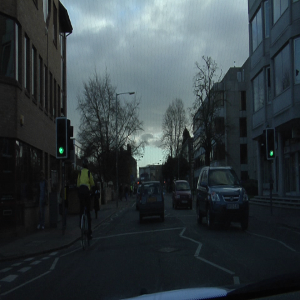}
	\end{subfigure}
    \begin{subfigure}{.30\linewidth}	
  	\includegraphics[width=\linewidth]{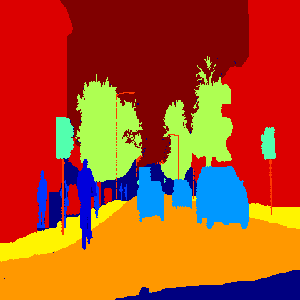}
	\end{subfigure}
    \begin{subfigure}{.30\linewidth}  	
  	\includegraphics[width=\linewidth]{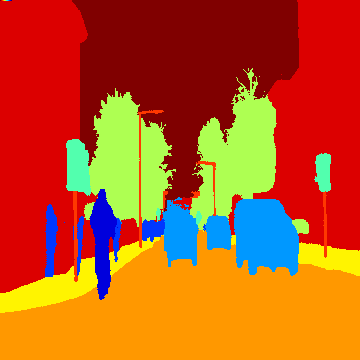}
  	\end{subfigure}
    }
    \\
    {
    \begin{subfigure}{.30\linewidth}
	\includegraphics[width=\linewidth]{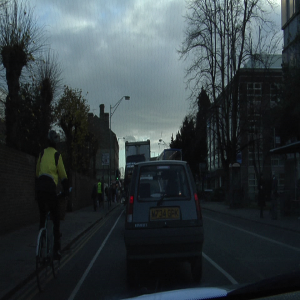}
	\end{subfigure}
    \begin{subfigure}{.30\linewidth}	
  	\includegraphics[width=\linewidth]{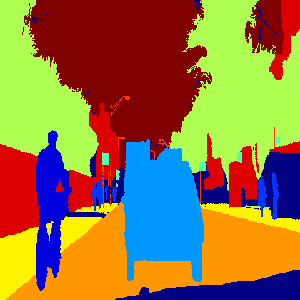}
	\end{subfigure}
    \begin{subfigure}{.30\linewidth}  	
  	\includegraphics[width=\linewidth]{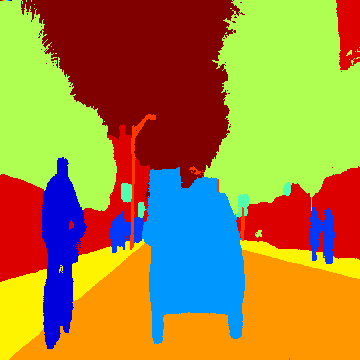}
  	\end{subfigure}
    }
    
    %\\
    %\subfloat{
  	%\includegraphics[width=.48\linewidth]{img/23734.pdf}
  	%\includegraphics[width=.48\linewidth]{img/23637.pdf}
    %}\\
    %\subfloat{
	%\includegraphics[width=.48\linewidth]{img/23648.pdf}
  	%\includegraphics[width=.48\linewidth]{img/24054.pdf}
    %}

    \caption{Qualitative results on the Camvid test set. The rows represent from left to right: Input image, ground truth and prediction of our approach. The first row exemplify a high quality segmentation with only a small inconsistency for the over-segmented tree class. The second and third rows show the higher performance obtained by DD-Net to fine classes, like pole, person and sign. The last row presents another prediction mistake, which is made between sky and vegetation, which happens in the top-right corner. The most common mistake of our approach is between the car hood and road. Such problems happens due to the reflection of the road image in the car hood.}
    \label{fig:camvid_qualitative}
  \end{figure*}

Qualitative results are shown in Figure \ref{fig:camvid_qualitative}. The first row shows examples in which the segmentation approach performs accurately, however, the tree class is over-segmented. The second and third rows show one of the strongest characteristics of our architecture, namely the highly detailed segmentation of challenging classes, such as person, pole and sign. The last row shows an example with less detailed structures and highlights the common confusion between sky and vegetation, which can be noticed in the fourth row, top-right side. The most common mistake of our approach is also presented in the last three rows, which is the confusion between the car hood and road. This is likely because the hood reflects the road image in most of the examples.

\subsection{Gatech dataset}

Gatech is a scene understanding dataset that consists of $63$ training videos and $38$ testing videos \parencite{Gatech}. It is much larger than Camvid, with $12000$ and $7000$ training and testing images respectively. However the annotations are often erroneous. Each video has between $50$ and $300$ frames, which are divided into $8$ classes: sky, ground, buildings, porous, humans, cars, vertical mix and main mix. One difference of evaluation metric is that given the dataset was originally designed to learn 3D geometric structure of outdoor video scenes the standard metric for this dataset is mean global accuracy.

\begin{figure}
    \centering
    {
    \begin{subfigure}{.32\linewidth}
       \includegraphics[width=\linewidth]{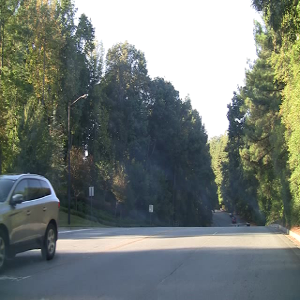}
    \end{subfigure}     
    \begin{subfigure}{.32\linewidth}    
       \includegraphics[width=\linewidth]{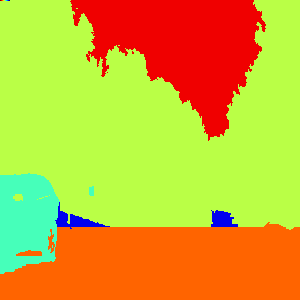}
    \end{subfigure}     
    \begin{subfigure}{.32\linewidth}          
        \includegraphics[width=\linewidth]{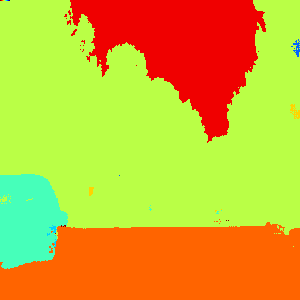}
    \end{subfigure}    
    }
    \\
    {
    \begin{subfigure}{.32\linewidth}
       \includegraphics[width=\linewidth]{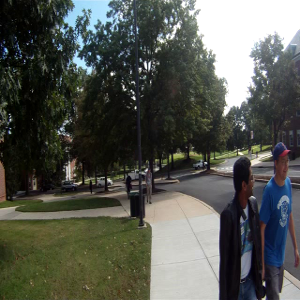}
    \end{subfigure}     
    \begin{subfigure}{.32\linewidth}    
       \includegraphics[width=\linewidth]{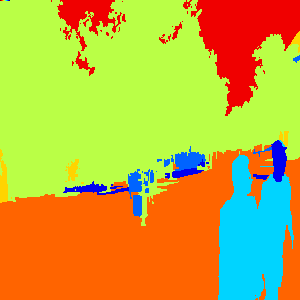}
    \end{subfigure}     
    \begin{subfigure}{.32\linewidth}          
        \includegraphics[width=\linewidth]{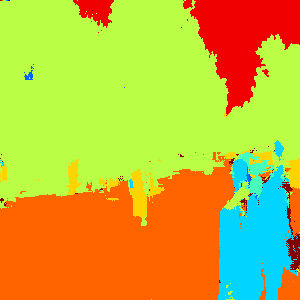}
    \end{subfigure}    
    }
    \\    
    {
    \begin{subfigure}{.32\linewidth}
       \includegraphics[width=\linewidth]{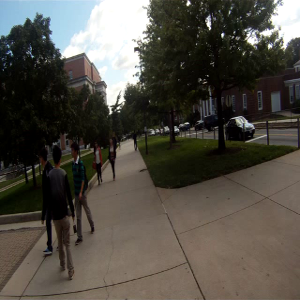}
    \end{subfigure}     
    \begin{subfigure}{.32\linewidth}    
       \includegraphics[width=\linewidth]{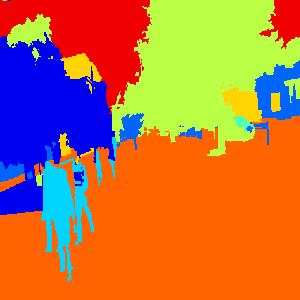}
    \end{subfigure}     
    \begin{subfigure}{.32\linewidth}          
        \includegraphics[width=\linewidth]{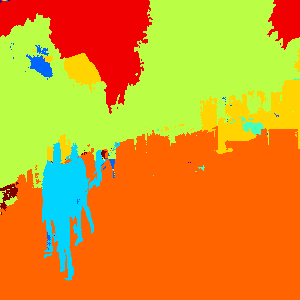}
    \end{subfigure}    
    }
    %\\
    %\subfloat{
  	%\includegraphics[width=.48\linewidth]{img/23734.pdf}
  	%\includegraphics[width=.48\linewidth]{img/23637.pdf}
    %}\\
    %\subfloat{
	%\includegraphics[width=.48\linewidth]{img/23648.pdf}
  	%\includegraphics[width=.48\linewidth]{img/24054.pdf}
    %}

    \caption{Qualitative results on the Gatech dataset. The rows represent from left to right: input image, ground truth and prediction of our approach. The first row shows good segmentation, with small false positive detection cases. The second and third examples count with poorly annotated masks that directly reflect on the segmentation predictions. For instance in the second row the right most person's head is wrongly annotated. Even with the presented problem our approach is still capable of segmenting classes like humans, ground, cars and sky correctly.}
    \label{fig:gatech_qualitative}
  \end{figure}

We pretrained our architecture on Cityscapes, as shown in Section \ref{pretraining}. We also provide results with training from scratch. In Table \ref{gatechTable}, we report the obtained results. As shown in the table, we outperform the compared methods not only in the fine-tuned scenario but also when the network is trained from scratch, even methods which include temporal consistency like \parencite{V2V} and \parencite{HDCNN} cannot produce as accurate results as the proposed technique. Figure \ref{fig:gatech_qualitative} presents qualitative results for the tested dataset, mainly showing that however with poorly annotated labels our technique can still produce good predictions.

\begin{table}[ht]
\centering
\caption{Results on Gatech dataset. The results are divided into trained from scratch and pre-trained. For both settings we outperform the current methods, even when compared to approaches which use temporal information.}
\label{gatechTable}
\begin{tabular}{l|c|c}
Method & Temporal Info & Acc. \\
\toprule
\toprule
2D-V2V-scratch {\scriptsize\parencite{V2V}}& No & $55.7$ \\
3D-V2V-scratch {\scriptsize\parencite{V2V}}& Yes & $66.7$ \\

DD-Net-scratch & No & $\bf{72.7}$ \\
\toprule

3D-V2V {\scriptsize\parencite{V2V}}& Yes & $76.0$\\
FC-DenseNet103 {\scriptsize\parencite{FCDense}} & No & $79.4$ \\
HDCNN {\scriptsize\parencite{HDCNN}} & Yes & $82.1$ \\
DD-Net & No & $\bf{83.1}$ \\

\bottomrule
\end{tabular}
\end{table}

\begin{table}
\centering
\caption{Results on Freiburg Forest dataset. Our proposed network outperform all the previous approaches, even when compared to a strong baselines which even use the same feature learning block at the encoder like DPDB-Net \parencite{oliveira2018icra}.
}
\label{forestDataset}
\begin{tabular}{l|c|c|c|c|c}
Method & \rot{Sky}  & \rot{Trail} & \rot{Grass} & \rot{Veg} & \rot{mIoU} \\
\toprule
\toprule
FCN  {\scriptsize \parencite{long_shelhamer_fcn}} & $-$ & $-$ & $-$ & $-$ & $77.0$\\
%SegNet {\scriptsize \parencite{segnet}} & $-$ & $-$ & $-$ & $-$ & $72.2$\\
ParseNet {\scriptsize \parencite{parsenet}} & $87.7$ & $81.8$ & $85.2$ & $85.2$ & $85.0$\\
E-Net {\scriptsize\parencite{ENet}} & $-$ & $-$ & $-$ & $-$ & $71.4$\\
M-Net {\scriptsize \parencite{Mnet}} & $89.2$ & $82.4$ & $84.9$ & $88.7$ & $86.3$ \\
Fast-Net {\scriptsize \parencite{FastNet}} & $90.4$  & $84.5$ & $86.7$ & $90.6$ & $88.0$ \\ 
GCN {\scriptsize\parencite{GCN}} & $91.9$ & $86.2$ & $86.4$ & $88.7$ & $88.3$ \\
DPDB {\scriptsize\parencite{oliveira2018icra}} & $92.3$  & $87.2$  & $87.8$ & $90.1$ & $89.4$ \\
\midrule
\midrule
%Ours-scratch & $92.64$  & $\bf{88.26}$  & $88.15$ & $90.18$ & $89.8$ \\
DD-Net & $\bf{92.9}$  & $\bf{88.9}$  & $\bf{88.5}$ & $\bf{90.7}$ & $\bf{90.2}$ \\
\bottomrule
\end{tabular}
\end{table}

\vspace{-0.3cm}

\subsection{Freiburg Forest dataset}

The Freiburg Forest dataset is an outdoor dataset for unstructured semantic segmentation \parencite{Valada2017}. Unstructured semantic understanding is critical for robots operating in real world scenarios. The dataset is composed by $325$ frames with pixel-level annotation, which has $203$ frames for training and $122$ frames for testing. The groundtruth is divided into five classes, such as sky, trail, grass, vegetation and obstacle.

%The experiments  presents an interesting characteristic that our method and the previous state-of-the-art share the same encoder, thus showing that the proposed approach can further improve a dataset which presents high performance results already.

\begin{figure}
    \centering
    {
    \begin{subfigure}{.32\linewidth}
       \includegraphics[width=\linewidth]{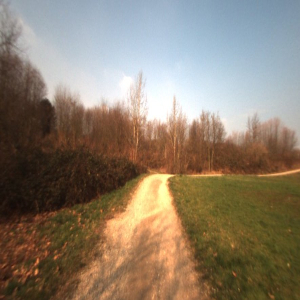}
    \end{subfigure}     
    \begin{subfigure}{.32\linewidth}    
       \includegraphics[width=\linewidth]{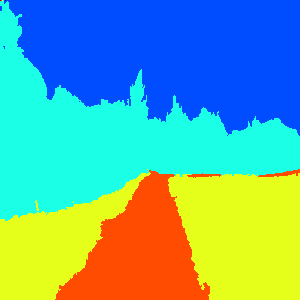}
    \end{subfigure}     
    \begin{subfigure}{.32\linewidth}          
        \includegraphics[width=\linewidth]{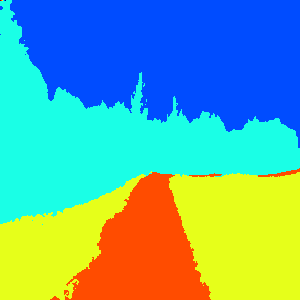}
    \end{subfigure}    
    }    \\
    {
    \begin{subfigure}{.32\linewidth}
       \includegraphics[width=\linewidth]{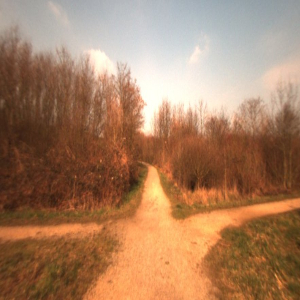}
    \end{subfigure}     
    \begin{subfigure}{.32\linewidth}    
       \includegraphics[width=\linewidth]{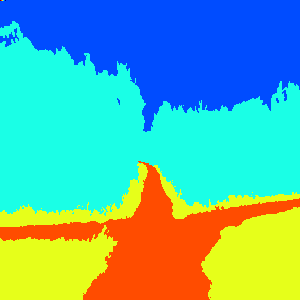}
    \end{subfigure}     
    \begin{subfigure}{.32\linewidth}          
        \includegraphics[width=\linewidth]{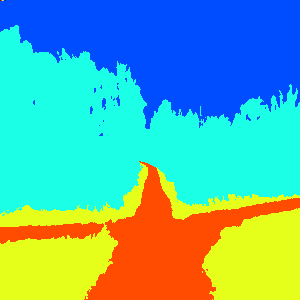}
    \end{subfigure}    
    }    \\
    {
    \begin{subfigure}{.32\linewidth}
       \includegraphics[width=\linewidth]{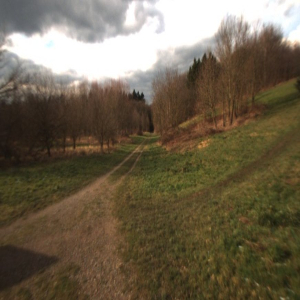}
    \end{subfigure}     
    \begin{subfigure}{.32\linewidth}    
       \includegraphics[width=\linewidth]{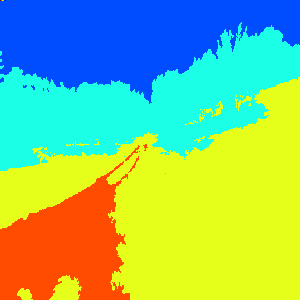}
    \end{subfigure}     
    \begin{subfigure}{.32\linewidth}          
        \includegraphics[width=\linewidth]{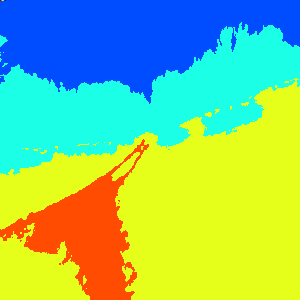}
    \end{subfigure}    
    }    \\    
    %\\
    %\subfloat{
  	%\includegraphics[width=.48\linewidth]{img/23734.pdf}
  	%\includegraphics[width=.48\linewidth]{img/23637.pdf}
    %}\\
    %\subfloat{
	%\includegraphics[width=.48\linewidth]{img/23648.pdf}
  	%\includegraphics[width=.48\linewidth]{img/24054.pdf}
    %}

    \caption{Qualitative results on the Freiburg Forest test set. The rows represent from left to right: Input image, ground truth and prediction of our approach. The first row depicts a good segmentation example with high quality segmentation mask. The second row exemplify a multi-path segmentation situation. The last row presents a sample of bad segmentation produced by our approach.}
    \label{fig:deepscene_qualitative}
  \end{figure}

Table \ref{forestDataset} reports the obtained state-of-the-art results and comparisons to the baseline. The experiment comprises an interesting comparison between our method and the previous state-of-the-art approach which uses an encoder with the same feature learning block. Such result displays that our decoder can further improve segmentation, even for a dataset which presents high performance results already. Another key aspect of our method, that is interesting for robotics applications, is the high IoU value for the traversable area of the dataset, namely trail class. High fidelity traversability prediction is of great interest for robotic path planning algorithms. Figure \ref{fig:deepscene_qualitative} presents multiple segmentation examples of our method for the Freiburg Forest dataset.

\subsection{Ablation Studies}
\label{subsec:ablation}

The performance of different components, like block learning, new blocks to encoder, decoder feature learning, decoder depth,  skip-connections, class balancing, network pre-training and deep decoders to modern networks will be presented in the following sections. All ablation studies until Section \ref{pretraining} are trained from scratch to provide a cleaner baseline comparison. We quantify the incremental performance of each component, which lead to our final approach. 

\subsubsection{Block learning analysis}

\begin{table*}[ht]
\centering
\caption{Ablation study with Dense and DPDB blocks. Results using the CamVid dataset. The experiment confirm the superior results of DPDB blocks for feature learning when compared to Dense Blocks in a single encoder-decoder setting. The experiment also quantify the gain of replacing a single decoder by our deep decoder approach.
}
\label{blocks}
\begin{tabular}{l||c|c|c|c|c|c|c|c|c|c|c||c|c}
Method & \rot{Building}  & \rot{Tree} &  \rot{Sky} &  \rot{Car} &  \rot{Sign} &  \rot{Road} &  \rot{Pedestrian} &  \rot{Fence} &  
\rot{Pole} &  \rot{Sidewalk} &  \rot{Cyclist} &  \rot{Global Acc.} & \rot{mIoU} \\
\toprule
\toprule
Dense blocks encoder-decoder &  $66.9$ & $62.4$ & $81.1$ & $45.5$ & $28.8$ & $83.2$ & $38.9$ & $26.8$ & 
$22.5$ & $62.0$ & $22.4$ & $81.0$ & $49.1$\\
DPDB encoder single decoder & $72.4$ &  $62.9$ & $88.6$ & $61.9$ & $30.0$ & $88.8$ & $44.8$ & $26.1$ & 
$23.6$ & $69.4$ & $33.1$ & $85.2$ & $54.7$\\
DD-Net & $\bf{80.4}$ & $\bf{74.0}$ & $\bf{92.1}$ & $\bf{78.8}$ & $\bf{41.3}$ & $\bf{94.5}$ & $\bf{56.9}$ & 
$\bf{25.7}$ & $\bf{34.1}$ & $\bf{80.7}$ & $\bf{61.8}$ & $\bf{90.4}$ & $\bf{65.5}$\\

\bottomrule
\end{tabular}
\end{table*}

\begin{table*}[ht]
\centering
\caption{Ablation study about the impact of feature learning after upsampling.  Results using the CamVid dataset. The experiment shows the importance of learning the upsampled features. Only upsampling displays the worst values. Standard upsampling followed by a convolution operation is outperformed by upsampling with a better feature learning approach, in our case upsampling followed by a Dense block.}
\label{ablationUpsampling}
\begin{tabular}{l||c|c|c|c|c|c|c|c|c|c|c||c|c}
Method & \rot{Building}  & \rot{Tree} &  \rot{Sky} &  \rot{Car} &  \rot{Sign} &  \rot{Road} &  \rot{Pedestrian} &  \rot{Fence} &  
\rot{Pole} &  \rot{Sidewalk} &  \rot{Cyclist} &  \rot{Global Acc.} & \rot{mIoU} \\
\toprule
\toprule
Only upsampling & $78.5$ & $72.8$ & $91.7$ & $75.3$ & $37.5$ & $93.7$ & $54.5$ & $22.7$ & $30.4$ & $78.9$ & $50.9$ & $89.5$ & 
$62.4$\\
Convolution after upsampling & $80.0$ &  $\bf{74.2}$ & $91.7$ & $77.0$ & $39.7$ & $93.1$ & $54.7$ & $22.3$ & $32.8$ & $77.7$ & 
$56.5$ & $89.9$ & $63.6$\\
DenseBlock after upsampling & $\bf{80.4}$ & $74.0$ & $\bf{92.1}$ & $\bf{78.8}$ & $\bf{41.3}$ & $\bf{94.5}$ & $\bf{56.9}$ & 
$\bf{25.7}$ & $\bf{34.1}$ & $\bf{80.7}$ & $\bf{61.8}$ & $\bf{90.4}$ & $\bf{65.5}$\\

\bottomrule
\end{tabular}
\end{table*}

An analysis of the choice between the deployment of DPDB blocks over dense blocks is presented in Table \ref{blocks}. We have tested three different settings, single encoder-decoder using only dense blocks, single encoder-decoder with DPDB blocks for the encoder and dense blocks for the decoder and our full approach with DPDB blocks for the encoder and dense blocks for the deep decoder. For single encoder-decoder approaches we can notice that using DPDB blocks can produce a gain superior to five percentage points over its dense block counter-part. While single encoder-decoder architectures perform well, the gap of performance between them and a multiple stage decoder, in our case superior to $10.8$ percentage points, makes deep decoders an interesting design choice for future semantic segmentation networks.

\subsubsection{New blocks to DD-Net}
\label{newblocks network}

Different learning blocks can incorporate new characteristics to segmentation. This setting is constituted by replacing the macro encoder by a new encoder with similar topology but with a different feature learning block. In the following experiments we used inverted residual blocks \parencite{MobileNetV2} and Dense blocks \parencite{DenseNet}. We choose inverted residual blocks based on its recent results on the MobileNet v2 architecture, which is a modern instance of efficient feature learning block. Additionally to inverted residual block we also use Dense blocks for further comparison. Table \ref{blocksDD} summarizes the results of the three tested configurations on the CamVid dataset.

\begin{table*}[ht]
\centering
\caption[Impact of learning blocks to DD-Nets]{Ablation study about the impact of learning blocks to DD-Nets.
}
\label{blocksDD}
\begin{tabular}{l||c|c|c|c|c|c|c|c|c|c|c||c|c}
Method & \rot{Building}  & \rot{Tree} &  \rot{Sky} &  \rot{Car} &  \rot{Sign} &  \rot{Road} &  \rot{Pedestrian} &  \rot{Fence} &  
\rot{Pole} &  \rot{Sidewalk} &  \rot{Cyclist} &  \rot{Global Acc.} & \rot{mIoU} \\
\toprule
\toprule
Inverted Residual + DD &  $77.8$ & $73.2$ & $91.2$ & $77.3$ & $36.0$ & $93.4$ & $49.6$ & $\mathbf{26.0}$ & $31.2$ & $77.8$ & $49.9$ & $89.3$ & $62.1$\\
Dense block + DD&  $78.9$ & $72.3$ & $91.5$ & $75.4$ & $37.5$ & $93.2$ & $53.6$ & $24.6$ & $30.9$ & $78.4$ & $53.6$ & $89.5$ & $62.7$ \\
DPDB block + DD & $\mathbf{80.4}$ & $\mathbf{74.0}$ & $\mathbf{92.1}$ & $\mathbf{78.8}$ & $\mathbf{41.3}$ & $\mathbf{94.5}$ & $\mathbf{56.9}$ & 
$25.7$ & $\mathbf{34.1}$ & $\mathbf{80.7}$ & $\mathbf{61.8}$ & $\mathbf{90.4}$ & $\mathbf{65.5}$\\
\bottomrule
\end{tabular}
\end{table*}

The results show that inverted residual blocks and dense blocks while well known efficient blocks are still not capable of surpassing our DPDB block in the same setting. Both inverted residual and dense blocks are performing around $62$ mean IoU percentage points, with a $0.6$ percent advantage to dense blocks. The only exception is the class fence where the inverted residual version is the highest IoU among the tested configurations. Our full approach with DPDB blocks at the decoder and deep decoders perform the highest with a mean IoU gain over the baselines by $2.8$ percentage points. This result again confirm the power of Dual-Path Dense-Blocks for feature learning in semantic segmentation tasks.

\subsubsection{Feature Learning at Decoder}
\label{featurelearingdecoder}

The impact of operations after upsampling features is explored in this section. We analyze the impact of no convolution after transposed convolution, convolution and dense-blocks after the upsampling operation. Table \ref{ablationUpsampling} summarized our obtained results where having a Dense Block after a transposed convolution can produce a gain of $3.1$ percentage points over only upsampling and a gain of $1.9$ percentage points over the standard strategy of including a convolution layer after the upsampling operation. The better feature representation given by Dense Blocks and the lack of residual connections which can potentially squash features, see Section \ref{analysis}, makes it the best candidate to learn upsampled feature at the decoder part.    

\begin{table*}[ht]
\centering
\caption{Decoder depth experiments on CamVid. The experiments indicate that,
with increasing number of decoder blocks, the model benefits from the deeper network.}
\label{DepthDecoder}
\begin{tabular}{l||c|c|c|c|c|c|c|c|c|c|c||c|c}
Method & \rot{Building}  & \rot{Tree} &  \rot{Sky} &  \rot{Car} &  \rot{Sign} &  \rot{Road} &  \rot{Pedestrian} &  \rot{Fence} &  
\rot{Pole} &  \rot{Sidewalk} &  \rot{Cyclist} &  \rot{Global Acc.} & \rot{mIoU} \\
\toprule
\toprule
One Decoder Block  & $80.3$ & $72.9$ & $91.8$ & $77.0$ & $40.2$ & $94.2$ & $52.9$ & $24.7$ & $32.8$ & $80.3$ & $53.3$ & $90.2$ 
& $63.7$\\
Two Decoder Blocks  & $\bf{80.9}$ & $73.7$ & $91.9$ & $78.1$ & $\bf{41.4}$ & $94.4$ & $55.1$ & $25.6$ & $33.7$ & $80.5$ & $58.2$ 
& 
$\bf{90.5}$ & $64.8$\\
Three Decoder Blocks & $80.4$ & $\bf{74.0}$ & $\bf{92.1}$ & $\bf{78.8}$ & $41.3$ & $\bf{94.5}$ & $\bf{56.9}$ & $\bf{25.7}$ & 
$\bf{34.1}$ & $\bf{80.7}$ & $\bf{61.8}$ & $90.4$ & $\bf{65.5}$\\

\bottomrule
\end{tabular}
\end{table*}

\begin{figure*}[ht]
    \centering
    {
    \begin{subfigure}{.19\linewidth}
       \includegraphics[width=\linewidth]{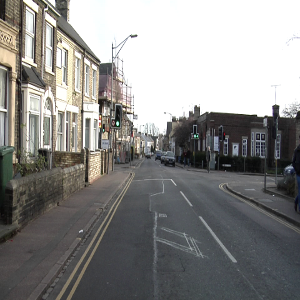}
    \end{subfigure}     
    \begin{subfigure}{.19\linewidth}    
       \includegraphics[width=\linewidth]{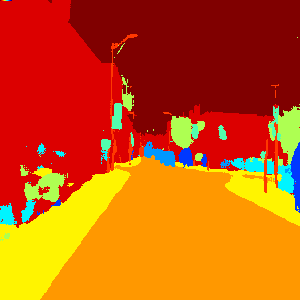}
    \end{subfigure}     
    \begin{subfigure}{.19\linewidth}          
        \includegraphics[width=\linewidth]{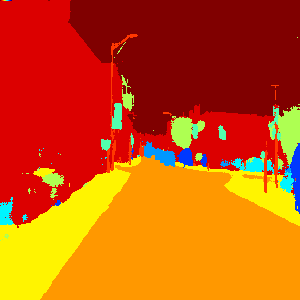}
    \end{subfigure}
    \begin{subfigure}{.19\linewidth}          
        \includegraphics[width=\linewidth]{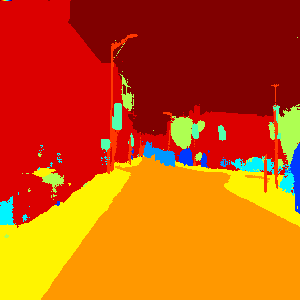}
    \end{subfigure}
    \begin{subfigure}{.19\linewidth}          
        \includegraphics[width=\linewidth]{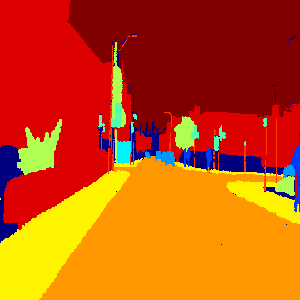}
    \end{subfigure}    
    }    \\        
    {
    \begin{subfigure}{.19\linewidth}
       \includegraphics[width=\linewidth]{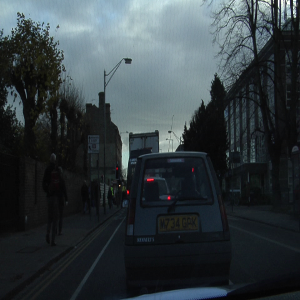}
    \end{subfigure}     
    \begin{subfigure}{.19\linewidth}    
       \includegraphics[width=\linewidth]{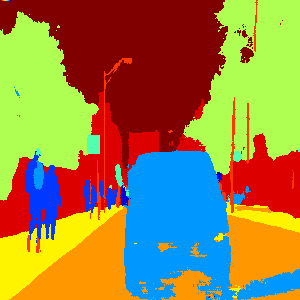}
    \end{subfigure}     
    \begin{subfigure}{.19\linewidth}          
        \includegraphics[width=\linewidth]{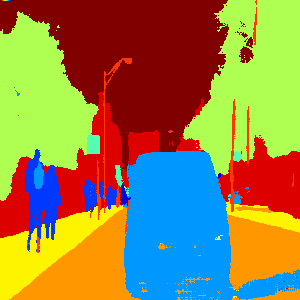}
    \end{subfigure}
    \begin{subfigure}{.19\linewidth}          
        \includegraphics[width=\linewidth]{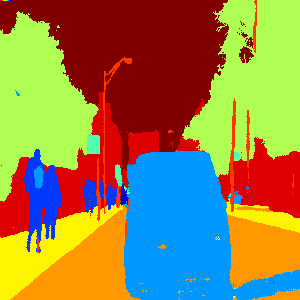}
    \end{subfigure}
    \begin{subfigure}{.19\linewidth}          
        \includegraphics[width=\linewidth]{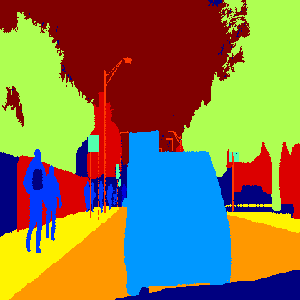}
    \end{subfigure}    
    }    \\ 
    {
    \begin{subfigure}{.19\linewidth}
       \includegraphics[width=\linewidth]{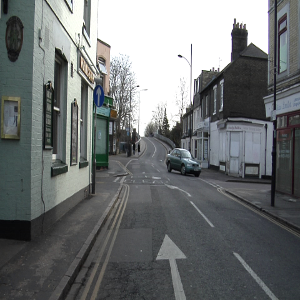}
    \end{subfigure}     
    \begin{subfigure}{.19\linewidth}    
       \includegraphics[width=\linewidth]{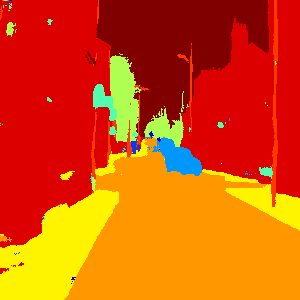}
    \end{subfigure}     
    \begin{subfigure}{.19\linewidth}          
        \includegraphics[width=\linewidth]{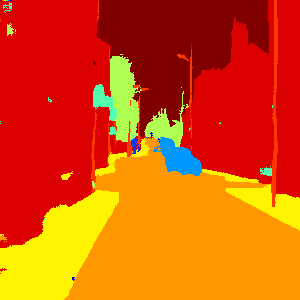}
    \end{subfigure}
    \begin{subfigure}{.19\linewidth}          
        \includegraphics[width=\linewidth]{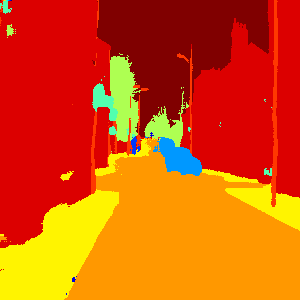}
    \end{subfigure}
    \begin{subfigure}{.19\linewidth}          
        \includegraphics[width=\linewidth]{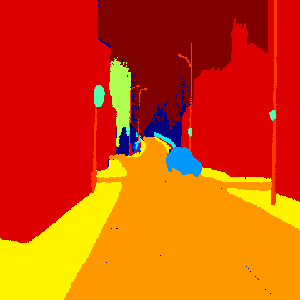}
    \end{subfigure}    
    }    \\ 
    
    \caption{Depth of decoder experiments. The columns show from left to right: Input image, result of the one-unit decoder, result of the two-unit decoder, results of the three-unit decoder and ground truth. What can be seen is the reduction of false-positive prediction with more units at the decoder. For instance the car class in the middle row presents gradually less holes.}
    \label{fig:depth_qualitative}
\end{figure*}

\subsubsection{Depth of Decoder}
\label{sec:decoderdepth}

The following experiments aim to understand the impact of the number of connected decoder-encoder units to the network's performance. The key feature we target to incorporate is that each included unit can capture more contextual information and produce higher fidelity predictions. To validate this goal, we trained multiple networks which gradually increase the number of units until we face memory limitations to further increment. The tested topologies are presented in Table \ref{DepthDecoder}. The observed results show a consistent performance improvement with the growth of the decoder-encoder units, respectively we got a increment of $1.1$ percentage points in the IoU metric from one to two decoders and $0.7$ percentage points from two to three units. The main gains are obtained by the least frequent and more complex classes, namely pedestrian and cyclist. Pedestrian exhibits an $4$ percentage points improvement while the class cyclist presents the largest improvement of $8.5$ percentage points between one and three units. Qualitative results with the different settings are presented in Figure \ref{fig:depth_qualitative}.

The results shown in Figure \ref{fig:depth_qualitative} mainly exemplify the false positive detection reduction with the decoder depth increment. The first column shows the specific case of the building class, the bottom left area presents multiple false positive regions, which are reduced with the increment of units. The second column is a sample of the better segmentation of the car class, holes are gradually closed, producing a boost of $1.8$ percentages points for this class. Column three is another example of the reduction on false-positive detections, where more units can reduce such occurrence.

\subsubsection{Skip-connections and resolution}

\begin{table*}[ht]
\centering
\caption{Ablation study on the impact of the different types of skip-connections on three different datasets. F stands for forward connection, B for backward, and R for residual. \textbf{Top:} CamVid dataset. \textbf{Bottom left:} Freiburg Forest dataset. \textbf{Bottom right:} Gatech dataset. The results support the claim that the proposed skip-connections provide a better context information flow which reflects on the quality of segmentation. The only drawback of the inclusion of backward and residual skip connections are in the case of corrupted annotations, for example presented in the Gatech dataset.} \label{skipconnections}

\begin{subtable}{\textwidth}
\sisetup{table-format=-1.2}   % 2 decimals, leave space for minus sign
\centering
   \begin{tabular}{l||c|c|c|c|c|c|c|c|c|c|c||c|c}
Method & \rot{Building}  & \rot{Tree} &  \rot{Sky} &  \rot{Car} &  \rot{Sign} &  \rot{Road} &  \rot{Pedestrian} &  \rot{Fence} &  
\rot{Pole} &  \rot{Sidewalk} &  \rot{Cyclist} &  \rot{Global Acc.} & \rot{mIoU} \\
\toprule
\toprule
F  & $79.7$ & $73.2$ & $91.6$ & $78.6$ & $41.2$ & $93.9$ & $54.1$ & $23.3$ & $30.8$ & $80.0$ & $56.4$ & $89.9$ & $63.9$\\
FB  & $\bf{81.0}$ & $73.6$ & $92.0$ & $78.0$ & $39.4$ & $93.8$ & $\bf{57.6}$ & $\bf{28.4}$ & $31.9$ & $80.1$ & $52.4$ & $90.3$ & $64.4$\\
FR & $80.7$ & $\bf{74.6}$ & $\bf{92.1}$ & $77.8$ & $39.9$ & $94.0$ & $55.6$ & $24.0$ & $\bf{34.6}$ & $80.0$ & $56.1$ & $\bf{90.4}$ & $64.5$\\
FBR & $80.4$ & $74.0$ & $\bf{92.1}$ & $\bf{78.8}$ & $\bf{41.3}$ & $\bf{94.5}$ & $56.9$ & $25.7$ & 
$34.1$ & $\bf{80.7}$ & $\bf{61.8}$ & $\bf{90.4}$ & $\bf{65.5}$\\
\bottomrule
\end{tabular}
   
\end{subtable}

\bigskip

\begin{tabular}{cc}
    \begin{minipage}{.5\textwidth}
        \centering

        \begin{tabular}{l|c|c|c|c|c}
Method & \rot{Sky}  & \rot{Trail} & \rot{Grass} & \rot{Veg} & \rot{mIoU} \\
        \toprule
        \toprule
        F & $92.7$ & $87.5$ & $87.7$ & $90.3$ & $89.56$\\
        FR & $92.6$  & $87.6$  & $87.9$ & $90.4$ & $89.65$ \\
        FB & $92.6$  & $87.6$  & $87.9$ & $90.5$ & $89.67$ \\
        FBR & $\bf{92.9}$  & $\bf{88.9}$  & $\bf{88.5}$ & $\bf{90.7}$ & $\bf{90.25}$ \\
\bottomrule
\end{tabular}
    \end{minipage} &

    \begin{minipage}{.5\textwidth}
    \centering
    \begin{tabular}{l|c}
        Method & Accuracy \\
        \toprule
        \toprule
        F  & $\bf{81.8}$\\
        FR & $80.9$ \\
        FB & $81.0$ \\
        FBR & $81.2$ \\
        \bottomrule
        \end{tabular}
    \end{minipage} 
\end{tabular}

\end{table*}

\begin{table*}[ht]
\centering
\caption{Experiments with varying resolution for our trained from scratch approach on the CamVid dataset. Higher resolution inputs are beneficial. }
\label{resolutions}
\begin{tabular}{l||c|c|c|c|c|c|c|c|c|c|c||c|c}
Method & \rot{Building}  & \rot{Tree} &  \rot{Sky} &  \rot{Car} &  \rot{Sign} &  \rot{Road} &  \rot{Pedestrian} &  \rot{Fence} &  
\rot{Pole} &  \rot{Sidewalk} &  \rot{Cyclist} &  \rot{Global Acc.} & \rot{mIoU} \\
\toprule
\toprule
DD-Net - $300\times300$ & $80.4$ & $74.0$ & $\bf{92.1}$ & $78.8$ & $41.3$ & $94.5$ & $56.9$ & $25.7$ & 
$34.1$ & $80.7$ & $61.8$ & $90.4$ & $65.5$\\
DD-Net - $360\times360$ & $\bf{81.5}$ & $\bf{75.1}$ & $\bf{92.1}$ & $\bf{80.1}$ & $\bf{43.1}$ & $\bf{95.0}$ & $\bf{58.8}$ & $\bf{28.3}$ & $\bf{35.4}$ & $\bf{82.2}$ & $\bf{58.1}$ & $\bf{91.0}$ & $\bf{66.3}$\\
\bottomrule
\end{tabular}
\end{table*}

This section is dedicated to quantify the effect of each type of skip connection and their combination. Additionally, we investigate the impact of resolution input to the network's segmentation accuracy. The first experiments quantify the impact of only forward skip-connections, with a combination of forward and backward, forward and residual and with our complete approach which is composed by forward, backward and residual connections. Table \ref{skipconnections} presents the obtained results. The experiment confirms the hypothesis that backward and residual connections can improve context information flow and consequently produce better segmentation masks. The FBR (Forward-Backward-Residual) model shows a considerable gain over the other settings. The Camvid and Freiburg Forest datasets also support this finding. Only on the Gatech dataset the inclusion of the backward and residual skip-connections has a negative impact in the network's segmentation quality metric. After verifying results qualitatively, we believe that the network actually performs better with the additional skip-connections (as on the other datasets), but suffers from the noisy annotation in the quality metric. Figure \ref{fig:gatech_qualitative} shows an example in the middle row, where DD-Net produces a consistent prediction to the person class, yet the (wrong) annotation assigns it to the background.

\begin{table*}[h]
\centering
\caption{Results with different weight functions for our trained from scratch model on the CamVid dataset. FL mean focal loss cross-entropy. Focal loss alone presented the lowest obtained result. Even when used together with our dynamic weights is still not as accurate as just using solely our approach. We believe the change in the priors induced by the focal loss function is harmful for multi-class learning.}
\label{weightFunctions}
\begin{tabular}{l||c|c|c|c|c|c|c|c|c|c|c||c|c}
Method &  \rot{Building}  & \rot{Tree} &  \rot{Sky} &  \rot{Car} &  \rot{Sign} &  \rot{Road} &  \rot{Pedestrian} &  \rot{Fence} & 
 
\rot{Pole} &  \rot{Sidewalk} &  \rot{Cyclist} &  \rot{Global Acc.} & \rot{mIoU} \\
\toprule
\toprule

No Class Balance  & $\bf{81.0}$ & $74.5$ & $92.4$ & $\bf{80.9}$ & $\bf{42.4}$ & $\bf{94.7}$ & $\bf{58.8}$ & $17.8$ & $31.5$ & 
$80.6$ & $54.6$ & $\bf{90.9}$ & $64.5$\\

Med. Freq. \parencite{Eigen15}  & $79.5$ & $\bf{74.7}$ & $\bf{92.5}$ & $79.3$ & $41.3$ & $93.8$ & $54.0$ & $22.8$ & 
$32.7$ & $78.8$ & $58.8$ & $90.2$ & $64.4$\\

FL \parencite{focalLoss}  & $79.2$ & $73.1$ & $\bf{92.2}$ & $81.1$ & $38.1$ & $93.6$ & $52.6$ & $24.2$ & $31.2$ & $78.9$ & $59.9$ & $89.2$ & $64.0$\\
\midrule
\midrule
FL - Dynamic Weights  & $80.3$ & $73.9$ & $92.2$ & $79.2$ & $39.6$ & $94.5$ & $55.3$ & $\bf{29.7}$ & $33.0$ & $80.4$ & $55.0$ & $\bf{90.9}$ & $64.8$\\

Dynamic Weights  & $80.4$ & $74.0$ & $92.1$ & $78.8$ & $41.3$ & $94.5$ & $56.9$ & $25.7$ & $\bf{34.1}$ 
& $\bf{80.7}$ & $\bf{61.8}$ & $90.4$ & $\bf{65.5}$\\

\bottomrule
\end{tabular}
\end{table*}

We further extend our experiments to test different resolutions. The resolutions include images from $300\times300$ to $360\times360$. Table \ref{resolutions} summarizes the results for both resolutions on CamVid dataset. Higher resolution inputs have the positive aspect of producing higher resolution segmentation masks at the end of the encoder and consequently a better defined initial segmentation prediction. As shown in our experiment increasing the resolution produces better results for all classes, outperforming all the previous tested settings.

\subsubsection{Class balance experiments}

\begin{table*}[ht]
\centering
\caption{Comparison of semantic segmentation performance on reduced version of the Cityscapes dataset. The reduced version aim to produce a CamVid like annotation pattern. As seen we outperform the previous state-of-the-art approach by more than eight percentage points, considering mean IoU. }
\label{cityscapes}
\begin{tabular}{l||c|c|c|c|c|c|c|c|c|c||c}
Method & \rot{Sky} &  \rot{Building} &  \rot{Road} &  \rot{Sidewalk} &  \rot{Cyclist} &  \rot{Vegetation} &  \rot{Pole} &  
\rot{car} &  \rot{Sign} &  \rot{Pedestrian} & \rot{mIoU} \\
\toprule
\toprule
FCN8s  \parencite{long_shelhamer_fcn}  & $81.20$ & $77.76$ & $92.80$ & $59.70$ & $49.23$ & $78.80$ & $21.5$ & $76.50$ & $48.84$ & $47.57$ &  $63.39$\\
SegNet \parencite{segnet}   & $69.93$ & $59.87$ & $83.25$ & $43.35$ & $27.25$ & $68.83$ & $19.23$ & $60.80$ & $23.81$ & $24.14$ &  $48.04$\\
ParseNet \parencite{parsenet}  & $82.21$ & $78.42$ & $92.76$ & $60.78$ & $50.30$ & $79.68$ & $22.86$ & $77.90$ & $49.12$ & $44.65$ &  $63.87$\\
AdapNet  \parencite{valada17icra}  & $\bf{87.13}$ & $83.14$ & $94.45$ & $68.93$ & $52.36$ & $85.72$ & $39.44$ & $84.16$ & $50.73$ & $47.81$ &  $69.39$\\
\midrule
\midrule
DD-Net   & $85.27$ & $\bf{88.99}$ & $\bf{97.22}$ & $\bf{80.06}$ & $\bf{67.64}$ & $\bf{90.32}$ & $\bf{52.50}$ & $\bf{91.08}$ & $\bf{60.94}$ & $\bf{68.95}$ & $\bf{78.30}$\\

\bottomrule
\end{tabular}
\end{table*}

Class imbalance is a natural characteristic of several segmentation datasets, causing two problems: (1) approach focus on more frequent classes that contribute to no useful learning signal of all classes; (2) easily classified classes can produce bias and degenerate models. Some approaches aim to solve such problem with median frequency class balance \parencite{Eigen15} or Focal Loss (FL) \parencite{focalLoss}. Median frequency class balance computes over the whole dataset a set of static weight classes given by $ac = median\_freq/freq(c)$. Class specific frequency $freq(c)$ is the total number of pixels in images where $c$ is present, and median frequency is the median of these frequencies. Focal loss is a dynamic scale weight function to the cross entropy loss. The aim is to reshape the loss function to down-weight easy examples and consequently focus training on hard examples. Focal loss cross entropy is designed to binary classification, however we implement an $\alpha$-balanced multi-class implementation based on the Equation \ref{focalloss}.

\begin{equation}
    F_L(p_t) = -\alpha_t\left ( 1 - p_t \right )^\gamma\log\left ( p_t \right )
    \label{focalloss}
\end{equation}

where $p_t \in [0, 1]$ is the model's estimated probability for the class $t$, $\alpha_t \in [0, 1]$ is the corresponding  weight factor and $\gamma \in [0,5]$ is the focusing parameter. The component $\left ( 1 - p_t \right )$ is the modulating factor for the cross entropy, which is responsible for adjusting the class weight.       
Additionally, we also tested our dynamic weight approach using focal loss, see Equation \ref{focallossDW}. This configuration aims to test how our area centric approach will behave with a decay function. The obtained results are presented in Table \ref{weightFunctions}. The results show that median frequency class balancing and no class balancing present similar results. For the focal loss experiment the results deteriorate when compared to the other settings, even when dynamic weight is included the obtained values are inferior to our sole weight function. The probable limitation of focal loss approaches is the aggressive change on the distribution of the values. Our approach, different from focal loss, does not change the class weights for the same input over time and consequently does not present such limitation.  

\begin{equation}
    F_L(p_t) = -\alpha_tDW_{bounded,t}\left ( 1 - p_t \right )^\gamma\log\left ( p_t \right )
    \label{focallossDW}
\end{equation}

\subsubsection{Network pre-training}
\label{pretraining}

Neural networks training usually requires large datasets with ground truth annotations. Data augmentation can alleviate this limitation through geometric and appearance transformation to the current dataset in order to produce more samples. However, presenting positive practical results it still cannot replace the need of thousands of labelled images for optimization. The following experiments will quantify the gains of network pre-training.

\begin{table*}[ht]
\centering
\caption{Pre-training results on CamVid dataset. The experiment presents a training from scratch vs pre-trained on Cityscapes version of our approach. As depicted our pre-trained model outperform the previous setting by $6.9$ mean IoU percentage points.}
\label{camvidFT}
\begin{tabular}{l||c|c|c|c|c|c|c|c|c|c|c||c|c}
Method & \rot{Building}  & \rot{Tree} &  \rot{Sky} &  \rot{Car} &  \rot{Sign} &  \rot{Road} &  \rot{Pedestrian} &  \rot{Fence} &  
\rot{Pole} &  \rot{Sidewalk} &  \rot{Cyclist} &  \rot{Global Acc.} & \rot{mIoU} \\
\toprule
\toprule
DD-Net - Scratch & $81.5$ & $75.1$ & $92.1$ & $80.1$ & $43.1$ & $95.0$ & $58.8$ & $28.3$ & $35.4$ & $82.2$ & $58.1$ & $91.0$ & $66.3$\\
DD-Net - Cityscapes FT   & $\bf{85.3}$ & $\bf{79.4}$ & $\bf{93.0}$ & $\bf{86.7}$ & $\bf{51.4}$ & $\bf{96.7}$ & $\bf{68.5}$ & $\bf{41.1}$ & $\bf{44.6}$ & $\bf{88.0}$ & $\bf{70.4}$ & $\bf{93.1}$ & $\bf{73.2}$\\

\bottomrule
\end{tabular}
\end{table*}

Network pre-training constitutes one of the main steps for CNN optimization. Given more diverse datasets, for instance Cityscapes \parencite{cityscapes} and Imagenet \parencite{imagenet_cvpr09}, fine-tuning on such datasets can further improve performance due to a richer feature representation. In automotive environments the Cityscapes dataset is a common dataset for such pre-training. It contains $2975$ training images and $500$ validation images. Cityscapes is a highly challenging benchmark, given the $50$ cities with different weather conditions, seasons and many dynamic objects. We adjusted the annotation to match the KITTI and Camvid datasets. Annotations for the modified Cityscapes are: sky, building, road, sidewalk, cyclist, vegetation, pole, car, sign and pedestrian. Table \ref{cityscapes} presents the obtained results training and testing on Cityscapes following the same setting as \parencite{valada17icra}. Our obtained results largely outperform all the compared results, even when compared to the AdapNet, which is a multi-resolution, multi-GPU, residual net based architecture.   

The promising results achieved on the Cityscapes dataset motivated the experiment of using the trained weights on other datasets, for example fune-tuning for the CamVid dataset. Table \ref{camvidFT} summarizes the results with and without fine-tuning (FT). Pre-training produces an improvement superior to $6$ percentage points in the intersection over union metric, also individually producing gain in all the CamVid classes. Additionally, we also experimented the impact of fine-tuning with different datasets for the Gatech setting. Table \ref{GatechFT} presents the impact of pre-training for the dataset, given the low number of samples the model using CamVid weights performs almost $2$ percentage points worse than the Cityscapes pre-trained counter-part. The diversity of the learned features from Cityscapes can produce a more generic model which consistently improve the results on all compared datasets.               

\begin{table}[ht]
\centering
\caption{Gatech results without fine-tuning and using CamVid and Cityscapes as pre-trained models. We obtain a gain superior to eight percentage points between training from scratch and using CamVid as initial weights. The presented improvement can be further extended, almost two percentage points, when pre-trained on Cityscapes.}
\label{GatechFT}
\begin{tabular}{l|c}
        Method & Accuracy \\
        \toprule
        \toprule
        Gatech only  & $72.7$\\
        CamVid+fine-tuning  & $81.2$\\
        Cityscapes+fine-tuning & $\bf{83.1}$ \\
        \bottomrule
        \end{tabular}
\end{table}

\subsubsection{Deep Decoder to modern networks}
\label{effect_decoders}

This study aims to measure the impact of deep decoders to modern techniques such as DeepLab v2 \parencite{DeeplabV2} and the DeepLab v3 family \parencite{DeepLabv3}. Our goal is to show if our approach can further improve state-of-the-art semantic segmentation methods.   

DeepLab v2 is an encoder-decoder architecture which make use of dilated convolution, also know as atrous convolution, to propose a new pyramidal model called Atrous Spatial Pyramid Pooling (ASPP). ASPPs when combined with a fully connected Conditional Random Field (CRF) post processing module can produce improved results. Additionally a multi-scale input technique is designed to further improve the network \parencite{Ponce2006}. The main contribution resides on the application of atrous convolutions to semantic segmentation, when compared to regular convolution with larger filters, atrous convolution allow to enlarge the field of view without increasing the amount of computation required.  

The experimental setting consists of taking Deeplab v2 and reporting results with and without our deep decoder, in order to measure the impact of our method to a standard benchmark architecture. Based on hardware limitations we were unable to use the full DeepLab v2 system, for that situation we use the same settings for both experiments, input resolution of $256 \times 512$, batch size of $2$, not use multi-scale inputs and no CRF post processing. For the reported results we are using the full cityscapes dataset and report the obtained results on the cityscapes validation set, see Table \ref{DecoderDeeplav2}.

\begin{table}[h!]
\centering
\caption[DeepLab v2 and DeepLab v2 with our deep decoder]{Comparison between the DeepLab v2 and DeepLab v2 with our deep decoder. Different from other experiments we are using the full cityscapes dataset and DeepLab v2 pre-trained on ImageNet.}
\label{DecoderDeeplav2}
\begin{tabular}{l|c}
        Method & mIoU \\
        \toprule
        \toprule
        DeepLab v2  & $65.6$\\
        DeepLab v2 with DeepDecoder  & $68.1$\\
        \bottomrule
        \end{tabular}
\end{table}

The changes to the baseline to include our deep decoder is the addition of forward skip connections between the encoder and the new decoder and a deep decoder with two decoders and one encoder, the full deep decoder is not possible to be implemented due to memory limitations. The Table \ref{DecoderDeeplav2} shows the results obtained after training for $100$ epochs and presents a gain of $2.5$ percentage points with the inclusion of our deep decoder, which indicates the power of our approach to improve strong baselines like the DeepLab v2 architecture.     

\begin{figure}[ht] 
	\centering
	\includegraphics[width=.98\linewidth]{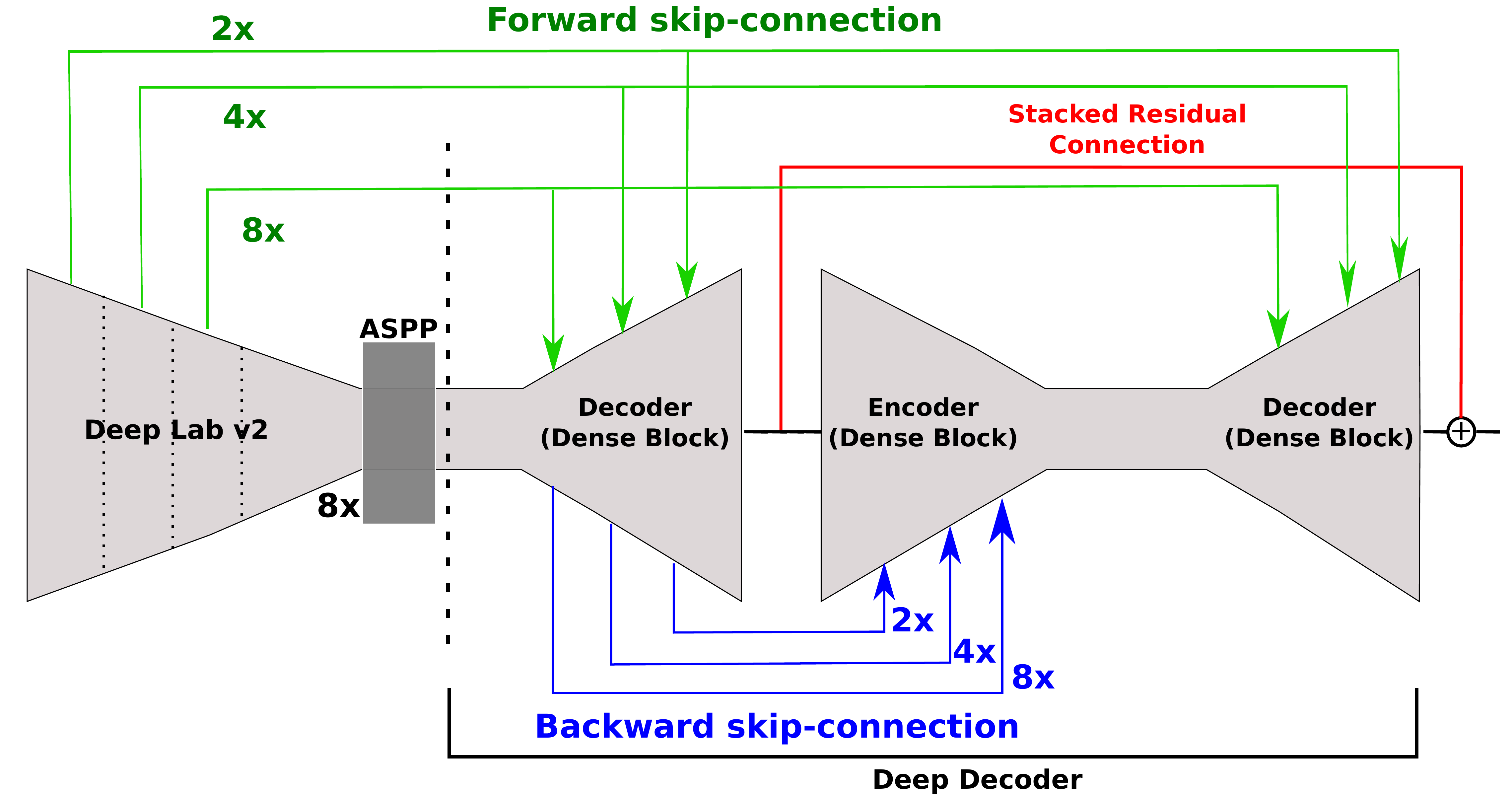}
	\caption[DeepLab v2 with deep decoder]{DeepLab v2 with deep decoder.} 
	\label{fig:deeplabv2DD}
\end{figure}

The modifications done to adequate DeepLab v2 to include our deep decoder are presented in Figure \ref{fig:deeplabv2DD}. The changes include the inclusion of three forward skip connections and the deep decoder module.

The next experiment consists of taking the state-of-the-art DeepLab v3+ model \parencite{DeepLabv3} and report results with and without our deep decoder. DeepLab v3+ is an evolution of DeepLab v3 where the authors proposed an improved version of their ASPP module and new decoder module to refine the segmentation results. 

Due to hardware limitations, we were unable to use the same resolution and batch size of the original paper, in this situation we used the same setting for both configurations. The input resolution of $256 \times 512$, batch size of $2$ and a single scale input approach was used. The reported results are obtained from training on the cityscapes validation set, Table \ref{DecoderDeeplav3}.

\begin{table}[ht]
\centering
\caption[DeepLab v3+ and DeepLab v3+ with our deep decoder]{Comparison between the DeepLab v3+ and DeepLab v3+ with our deep decoder. Different from other experiments we are using the full cityscapes dataset and DeepLab v3+ pre-trained on ImageNet.}
\label{DecoderDeeplav3}
\begin{tabular}{l|c}
        Method & mIoU \\
        \toprule
        \toprule
        DeepLab v3+  & $67.1$\\
        DeepLab v3+ with DeepDecoder  & $68.7$\\
        \bottomrule
        \end{tabular}
\end{table}

The changes to the baseline to include our deep decoder is the addition of one forward skip connection, at the same way it was implemented originally in the paper. The Table \ref{DecoderDeeplav3} shows the results obtained after training for $100$ epochs and presents a gain of $1.6$ percentage points with the inclusion of our deep decoder, which confirm the power and generality of our approach to improve strong baselines like the DeepLab v3+ architecture.

\begin{figure}[h] 
	\centering
	\includegraphics[width=.98\linewidth]{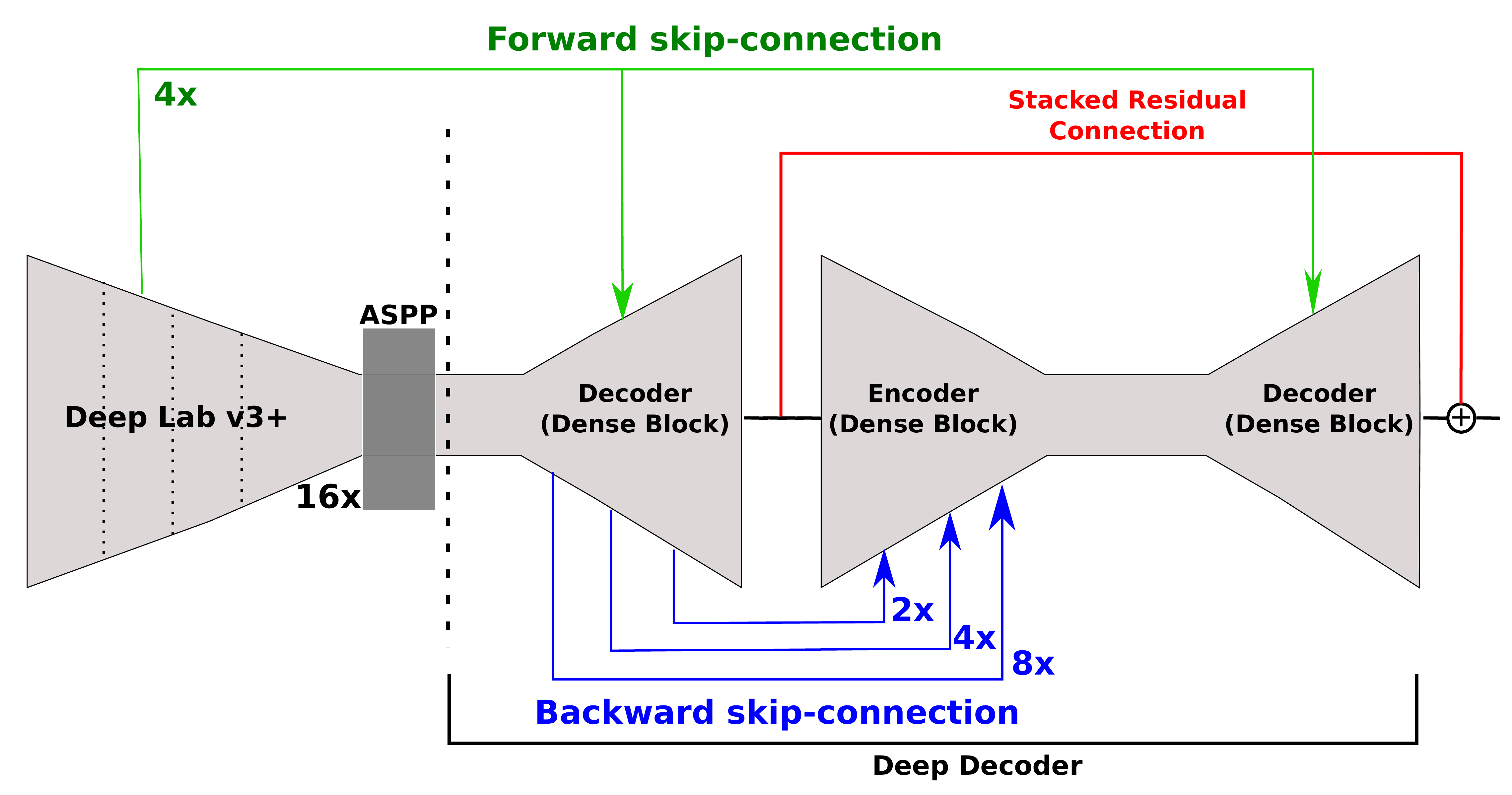}
	\caption[Deeplab v3+ with deep decoder]{Deeplab v3+ with deep decoder.} 
	\label{fig:deeplabv3DD}
\end{figure}

The modifications to include our deep decoder to the DeepLab v3+ model are presented in Figure \ref{fig:deeplabv3DD}. The changes only include the replacement of the DeepLab v3+ decoder module by our deep decoder approach. Table \ref{cityscapesTableFull} presents the IoU values for each class for the DeepLab v2 and DeepLab v3+ settings. The comparison between DeepLab v2 and the version with the inclusion of the proposed decoder shows the advantage of deep decoders for segmentation. Our approach not only shows a gain in the mean IoU of $2.5$ percentage points but also is consistently better for most classes, the only exceptions are the bus and motorcycle classes. Results on DeepLab v3+ quantifying the gain of our approach on the baseline architecture are shown next. The mean IoU gain is around $1.6$ percentage points, however some some classes like bus, truck and train we outperform the baseline by $10$, $11$ and $6$ percentage points respectively.     

We can conclude from this study that deep decoders shows improvement to semantic segmentation existing techniques. The gains will depend on the topology of the baseline approach, however even for strong baselines like DeepLab v3+ deep decoder are beneficial.

\begin{table*}[!htbp]
\centering
\scriptsize
\setlength\tabcolsep{5.0pt}
\caption[Results on Cityscapes dataset]{Results on Cityscapes dataset. Our deep decoder when combined with both approaches provided subtantial gains. In the case of deeplab v2 with DD we only had a lower score than our baseline for the bus class all the other classes we obtain better IoU scores, the only exception is motorcycle where we perform the same. For the comparison with deeplab v3+ our decoder provided a gain over the mean IoU and subtantial gains for the class bus, truck and train, with $10$, $11$ and $6$ percentage points respectively.}
\label{cityscapesTableFull}
%@{\hspace{0.8\tabcolsep}}
\begin{tabular}{l|c|c|c|c|c|c|c|c|c|c|c|c|c|c|c|c|c|c|c|c}
Method & \rot{Road} &  \rot{sidewalk}  & \rot{Building} & \rot{Wall} &  \rot{Fence} &  \rot{Pole} &  \rot{Traffic light} & \rot{Traffic Sign} &  \rot{Vegetation} &  \rot{Terrain} &  \rot{Sky} &  \rot{Person} & \rot{Rider} & \rot{Car} & \rot{Truck} & \rot{Bus} & \rot{Train} & \rot{Motorcycle} & \rot{Bicycle} & \rot{mIoU} \\
\toprule
\toprule
DeepLab v2 {\parencite{DeeplabV2}} & $96.$ & $74.$ & $89.$ & $50.$ & $42.$ & $44.$ & $49.$ & $62.$ & $90.$ & $54.$ & $92.$ & $71.$ & $49.$ & $91.$ & $65.$ & $\mathbf{72.}$& $45.$ & $46.$ & $66.$ & $65.6$\\
DeepLab v2 with DD & $\mathbf{97.}$ & $\mathbf{79.}$ & $\mathbf{90.}$ & $\mathbf{52.}$ & $\mathbf{47.}$ & $\mathbf{53.}$ & $\mathbf{54.}$ & $\mathbf{67.}$ & $\mathbf{91.}$ & $\mathbf{62.}$ & $\mathbf{94.}$ & $\mathbf{75.}$ & $\mathbf{52.}$ & $\mathbf{93.}$ & $\mathbf{73.}$& $70.$& $\mathbf{32.}$ & $46.$ & $\mathbf{69.}$ & $\mathbf{68.1}$\\
\midrule
\midrule
DeepLab v3+ {\parencite{DeepLabv3}} & $97.$ & $77.$ & $89.$ & $50.$ & $48.$ & $47.$ & $\mathbf{47.}$ & $61.$ & $90.$ & $\mathbf{59.}$ & $92.$ & $71.$ & $\mathbf{50.}$ & $92.$ & $64.$ & $65.$& $62.$& $46.$ & $66.$ & $67.1$\\
DeepLab v3+ with DD & $97.$ & $77.$ & $\mathbf{90.}$ & $\mathbf{51.}$ & $\mathbf{49.}$ & $\mathbf{52.}$ & $45.$ & $\mathbf{63.}$ & $\mathbf{91.}$ & $56.$ & $\mathbf{93.}$ & $71.$ & $46.$ & $\mathbf{93.}$ & $\mathbf{75.}$ & $\mathbf{77.}$& $\mathbf{68.}$& $\mathbf{47.}$ & $\mathbf{67.}$ & $\mathbf{68.7}$ \\
\bottomrule
\end{tabular}
\end{table*}

\subsubsection{Performance Tests}
\label{performance}

%Since the runtime depends much on the hardware and resolution, we present 
%results on various GPUs and at different resolutions for the Fast-Net 
%architecture.
%We tested seven different desktop GPUs and two mobile ones; see Table 
%\ref{KITTI_GPU}.
%Even on older GPUs, such as the GTX 680, the architecture achieves more than
%$10$ frames per second and fits into the GPU memory. The tests also reveal that the network is as fast on a GTX 980, GTX TITAN X and on the recently 
%release GTX 1070. 

Deployment runtime is a key aspect of methods which aim to be used into real life or mobile applications. The vast amount of possible hardware options can produce large performance differences. We aim with this study to benchmark our approach on various various GPUs, namely TITAN X, TITAN Xp, GTX 1080 and 180Ti, P100 and TITAN V. We tested five different modern GPUs, see Table \ref{GpusDDNet}.  

We can notice from this experiment that our approach can be deployed on GPUs with multiple memory sizes, ranging from $8$ gigabytes for the GTX 1080 to the $16$ gigabytes for the P100 GPUs. The top performing results are obtained on the TITAN V and 1080Ti cards. TITAN V cards can perform a full forward pass in less than $140$ milliseconds, constituting a $7$ frames per second capability. The slowest setting was using a TITAN X GPU in which a forward pass takes $306$ milliseconds on average.     

\begin{table}[h!]
\begin{center}
\caption[Runtime study for DD-Net]{Runtime depending on the GPU DD-Net.}
\label{KITTI_GPU}
\begin{tabular}{c|c}
 {\textbf{GPU}} & {\textbf{Forward Pass Time (ms)}}\\ \hline 
 GTX TITAN X & 306 \\
 GTX 1080 & 272 \\
 GTX TITAN Xp & 278 \\
 GTX 1080Ti & 201 \\
 P100 & 210 \\
 GTX TITAN V & 139 \\
 
 \end{tabular}
\label{GpusDDNet}
\end{center}
\end{table}

\begin{table}[h]
\begin{center}
\setlength\tabcolsep{3.5pt}
\caption[Experiment optimizing DD-Net for runtime]{Experiment optimizing DD-Net for runtime performance. Our full DD-Net can process $7$ frames per second, while changing the encoder block can make DD-Net to process $14$ frames per second. Further reducing the deep decoder to include only two decoder blocks makes our approach capable of processing $21$ frames per second.}
\label{KITTI_GPU_2}
\begin{tabular}{c|c|c}
 \textbf{Setting} & \textbf{mIoU} & {\textbf{Forward Pass Time (ms)}}\\ \hline
 DD-Net & $\mathbf{65.5}$ & $139$ \\
 Inverted residual with DD & $62.1$ & $71$ \\
 Inverted residual with slim DD & $61.4$ & $\mathbf{47}$ \\
 
\end{tabular}
\label{performanceOptimized}
\end{center}
\end{table}

Additionally to the full DD-Net approach, we also experimented focusing on optimizing the runtime of our approach. The first change consists of replacing the DPDB block by inverted residuals in the encoder, however we kept the full deep decoder. The results are obtained using the CamVid dataset and summarized in Table \ref{performanceOptimized}. The change between DPDB and inverted residual blocks reduce the need time for a single forward pass by half, nonetheless the mean IoU metric is $3.4$ percentage points lower. Such configuration can be important for the deployment of DD-net on robots with low computation capabilities. The new configuration can already provide a $14$ frames per second response time. Changing the encoder block gives a substantial saving, related to runtime, but we also experimented reducing the deep decoder size. Following the results from Section \ref{sec:decoderdepth} we reduced the size of our deep decoder from three decoder blocks to two decoder blocks. This modification impacted the overall performance by $4.1$ percentage points when compared to the full approach, although our technique now only needs $47$ milliseconds for a single forward pass, which constitutes a gain of almost three times when compared to our full approach and as consequence a method capable of meeting mobile robots requirements.

\subsection{Optical Flow Augmentation for Semantic Segmentation}

\begin{figure}[h!]
   \centering
    {
    \begin{subfigure}{.48\linewidth}
       \includegraphics[width=\linewidth]{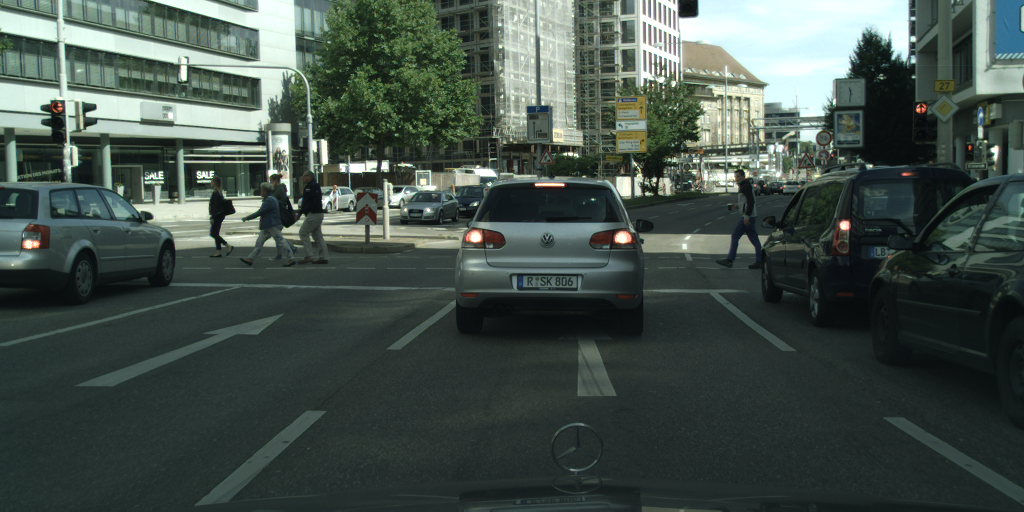}
    \end{subfigure}     
    \begin{subfigure}{.48\linewidth}    
       \includegraphics[width=\linewidth]{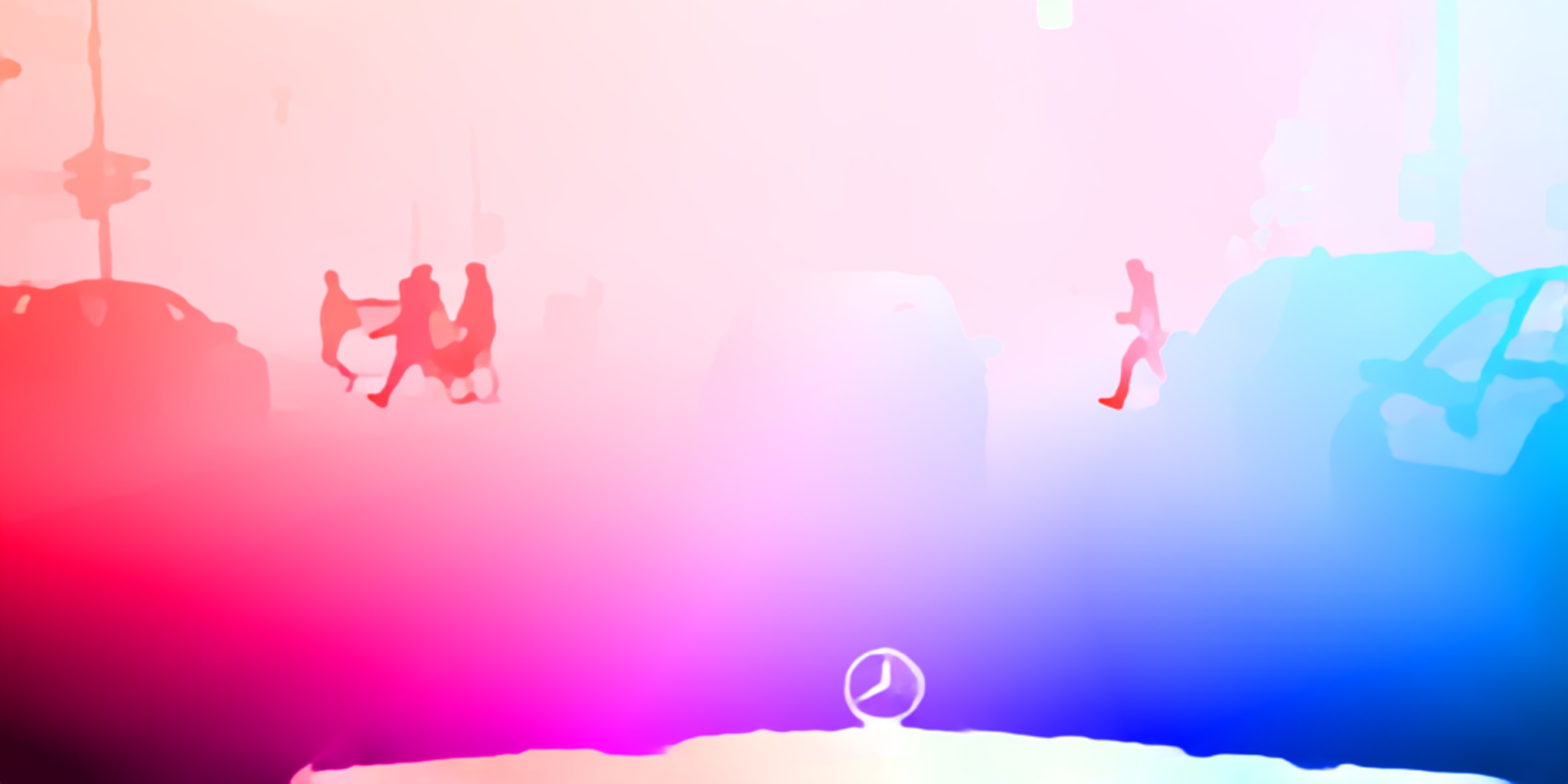}
    \end{subfigure}     
    }    \\    
    {
    \begin{subfigure}{.48\linewidth}
       \includegraphics[width=\linewidth]{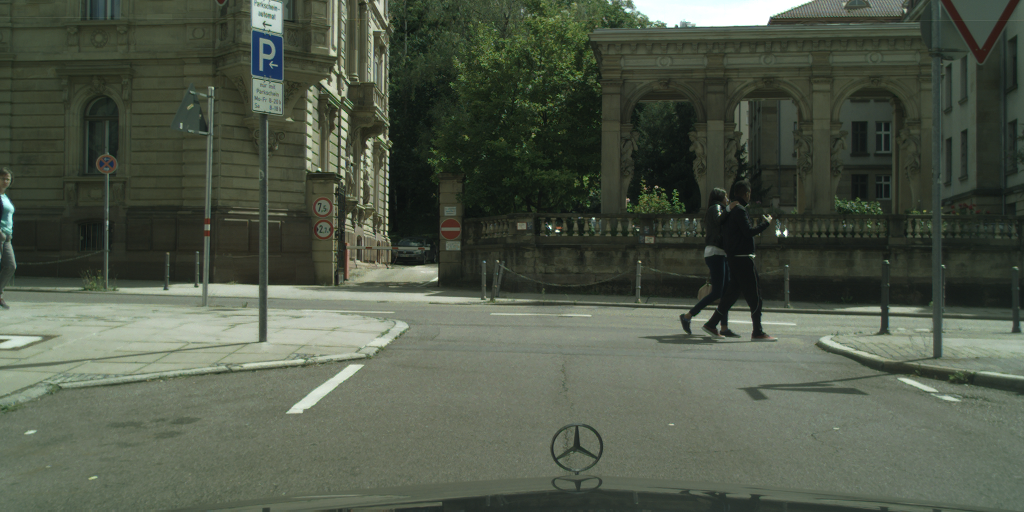}
    \end{subfigure}     
    \begin{subfigure}{.48\linewidth}    
       \includegraphics[width=\linewidth]{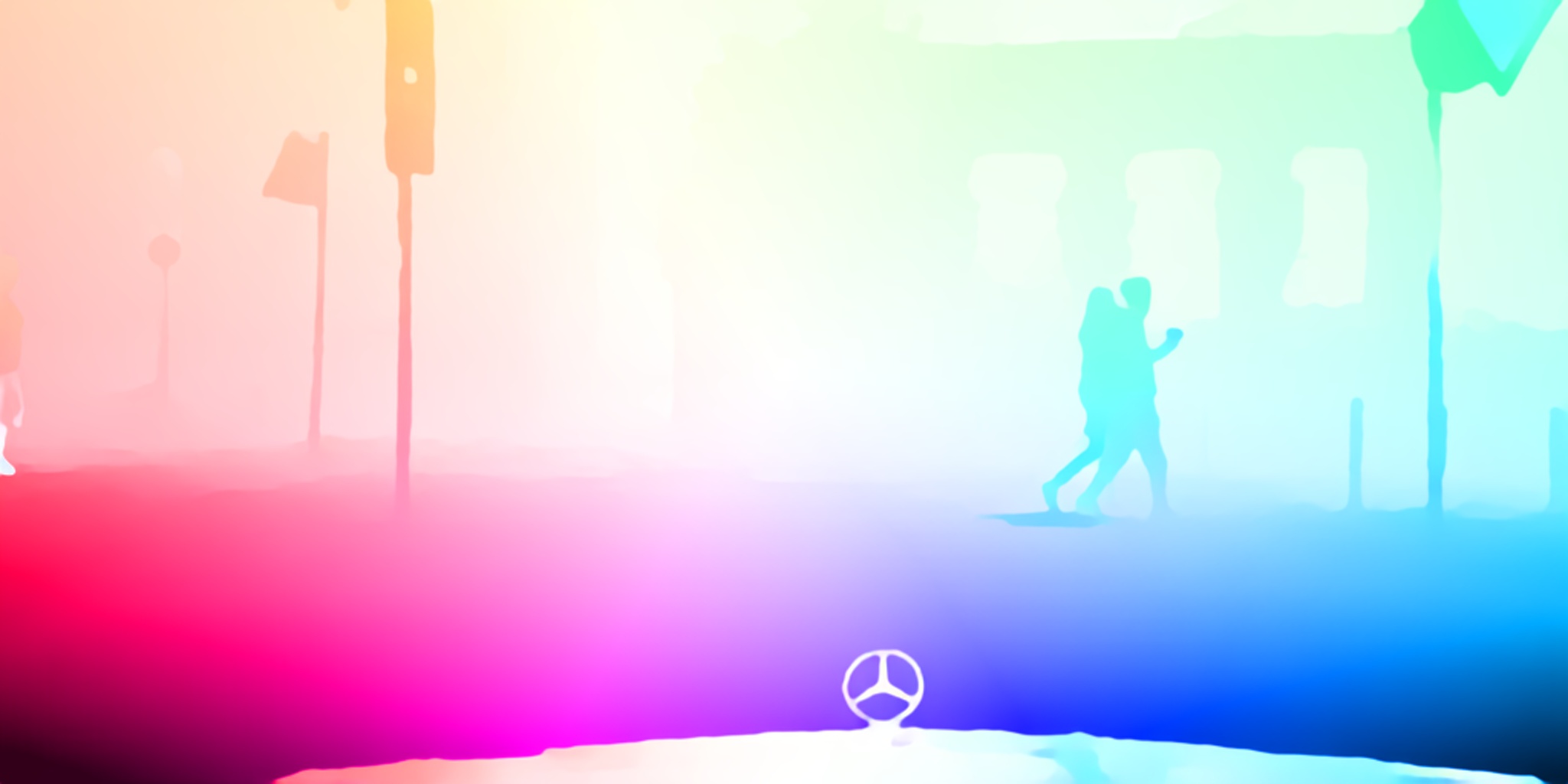}
    \end{subfigure}     
    }    \\ 
    {
    \begin{subfigure}{.48\linewidth}
       \includegraphics[width=\linewidth]{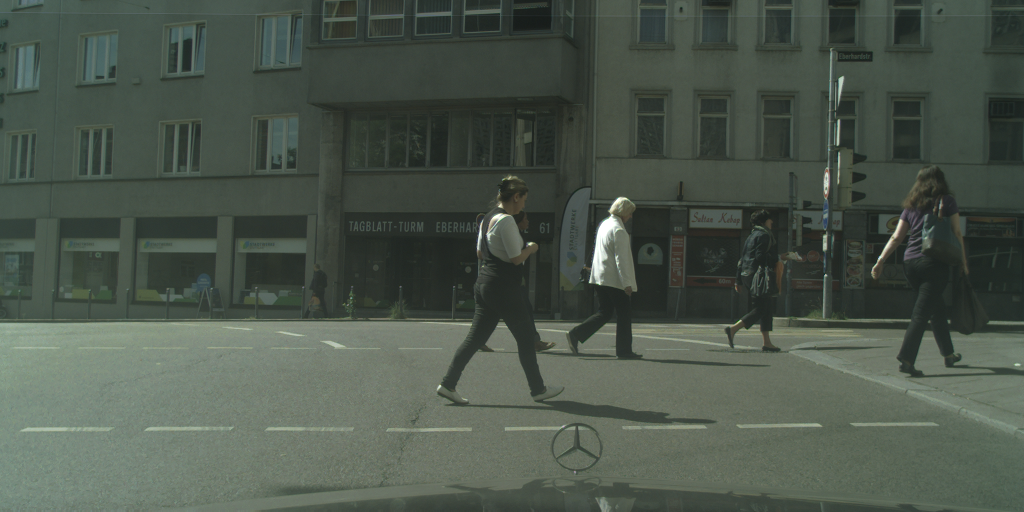}
    \end{subfigure}     
    \begin{subfigure}{.48\linewidth}    
       \includegraphics[width=\linewidth]{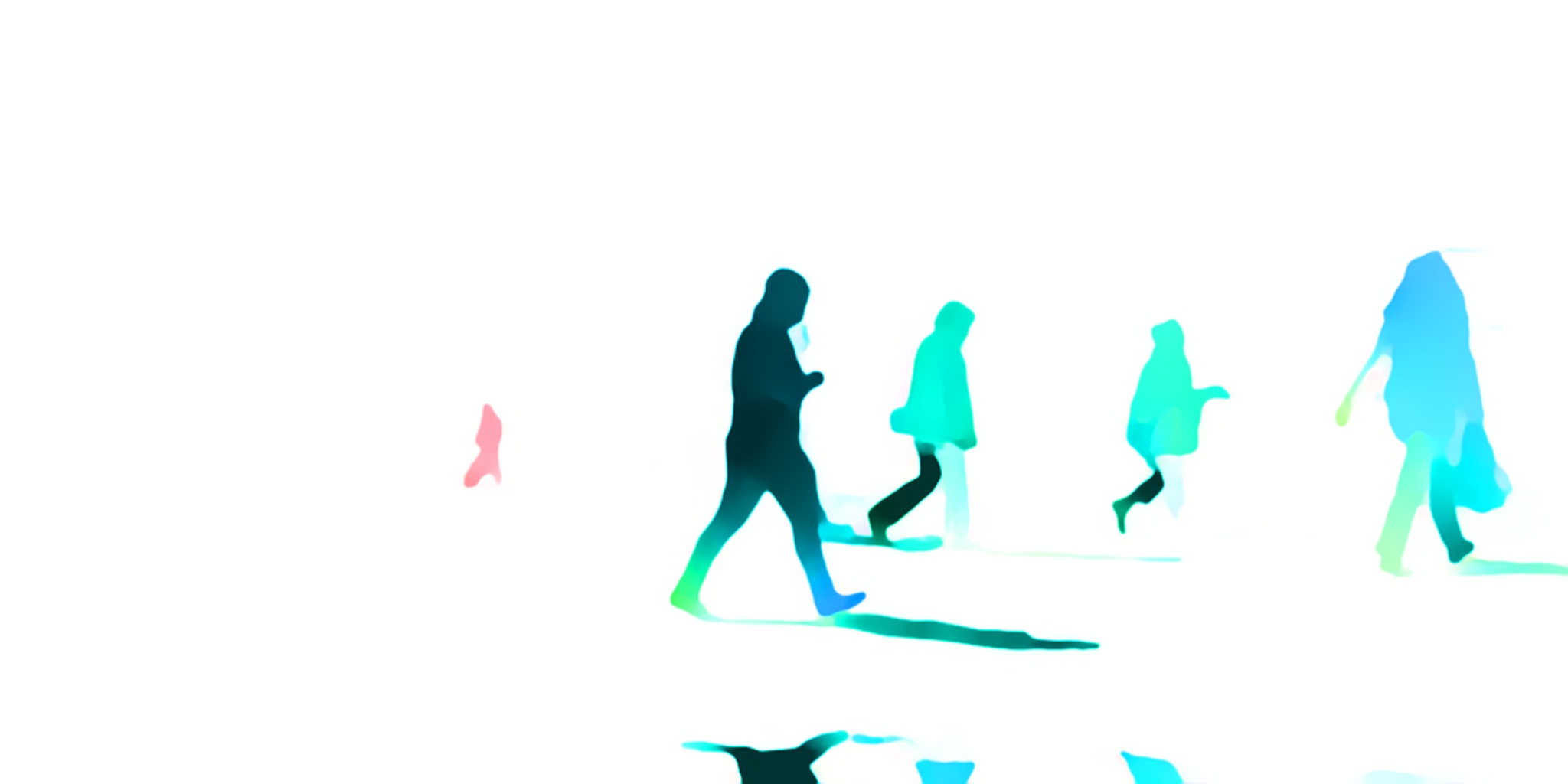}
    \end{subfigure}     
    }    \\ 
    {
    \begin{subfigure}{.48\linewidth}
       \includegraphics[width=\linewidth]{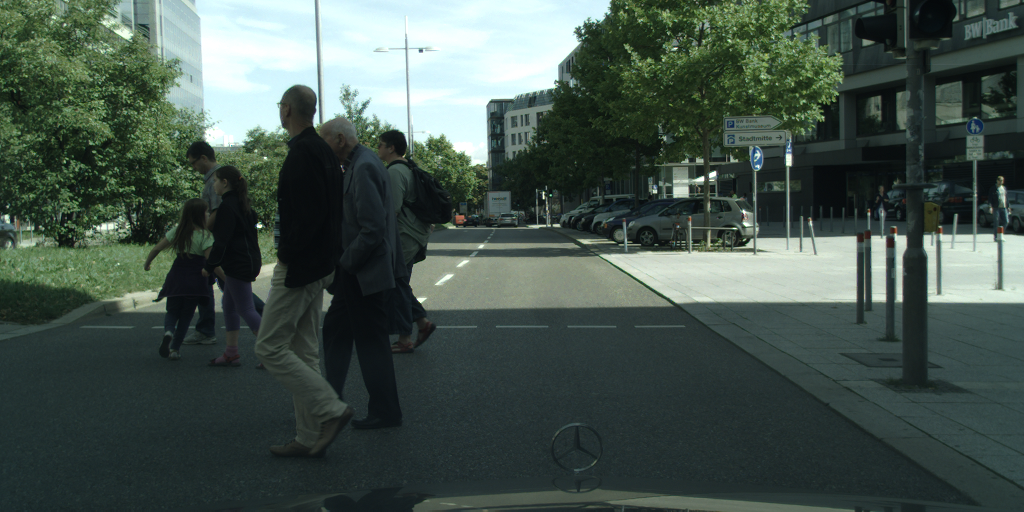}
    \end{subfigure}     
    \begin{subfigure}{.48\linewidth}    
       \includegraphics[width=\linewidth]{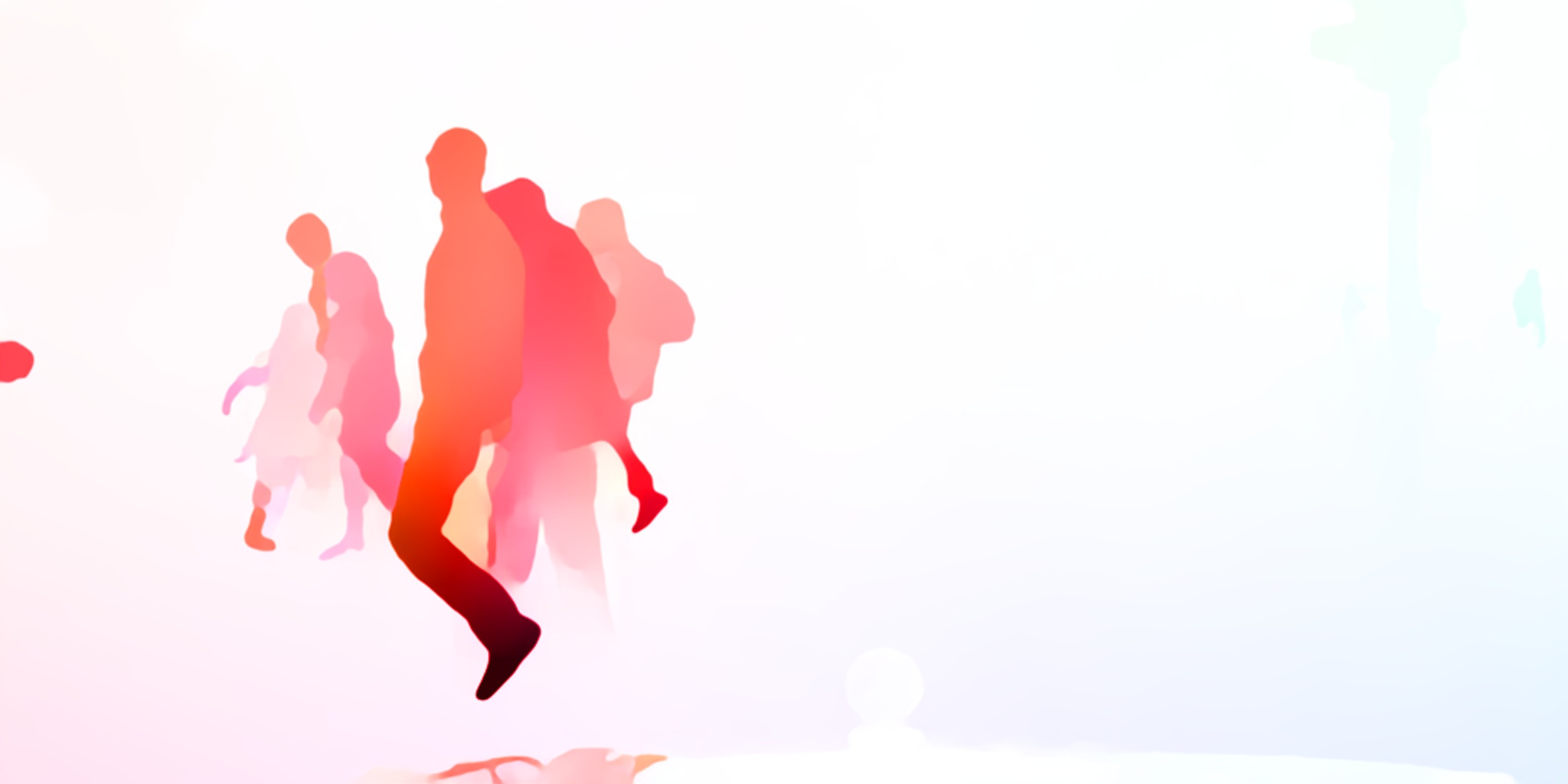}
    \end{subfigure}     
    }    
	\caption[Image and corresponding flow inputs]{Image and corresponding flow inputs. The top two rows are examples of optical flow with ego motion flow and independent object motion. The last bottom rows are examples without ego motion on the scene and for consequence easier instances for motion prediction.}
	\label{fig:flow_inputs}
\end{figure}

The majority of semantic segmentation methods rely only on appearance cues and don't exploit other input modalities. Some attempts to include depth were explored recently by \parencite{MousavianPK16, Zhang_2018_ECCV}, nonetheless motion clues have been less explored as an important clue for segmentation. Motion cues can be a challenging task because of the camera motion along with the motion of independent objects, see Figure \ref{fig:flow_inputs}.

Semantic segmentation approaches can benefit from the inclusion of motion clues, optical flow can provide
complementary cues about a dynamic scene that can be used to generate richer model of the scene. Additionally  motion clues can also be used for semantic motion segmentation, which is the ability of semantic segmentation to classify pixels as dynamic or static.

\begin{figure*}[ht!] 
	\centering
	\includegraphics[width=1.0\textwidth]{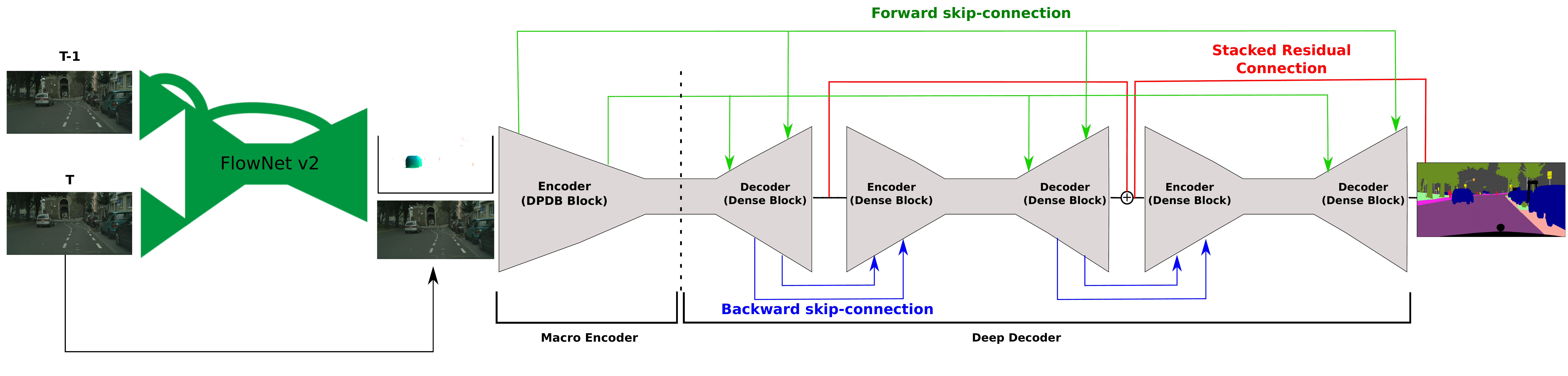}
	\caption[DD-Net Architecture with flow augmentation]{DD-Net Architecture with flow augmentation. For visual simplification we do not show the full flow network, in our case FlowNet v2. To obtain optical flow estimation for each image in the dataset, we used the FlowNet v2 to predict the flow map between each image and its previous frame. After it the magnitude and direction are channel fused to the RGB image and given to our DDNet.} 
	\label{fig:DDFLow}
\end{figure*}

Attempts to fuse appearance and motion clues have been proposed by \parencite{Hur2016JointOF, fusionseg, Vertens17iros}. The work from \parencite{Vertens17iros} is a method which uses flow information for semantic motion segmentation, while presenting good results its massive architecture, in terms of memory and gpu requirements, makes it unfeasible to robotics applications. \parencite{fusionseg} introduced a method which fuses appearance and motion for  agnostic foreground object segmentation. Another application of flow augmentation is to provide temporal consistent semantic segmentation, which is explored by \parencite{Hur2016JointOF}. Our network is conceptually closer to \parencite{Vertens17iros}, nevertheless we present a complete new design for the encoder and specially for the decoder. The proposed approach aims to use motion as a complementary cue to color. An overview of the proposed network is presented in Figure \ref{fig:DDFLow}. To obtain optical flow estimation for each image in the dataset, we used the FlowNet v2 model \parencite{flownet2} to predict the flow map between each image and its previous frame.

The flow augmentation takes two consecutive images and computes the flow. The flow representation used is the magnitude and direction in $2$ channels. We choose early fusion as modality fusion technique based on the low computational impact. Late fusion with single encoder streams for RGB and flow are prohibited in terms of computational requirement and computation overhead. The input for the DD-Net segmentation is the RGB + magnitude and direction. The remaining network is our deep decoder network.

Our experiments will focus on the Cityscapes dataset. The dataset includes the previous image frame relative to each annotated image in a separate subset of video sequences, which we use to obtain optical flow predictions. We report results on the compressed version of cityscapes with $10$ semantic classes. Table \ref{cityscapes_flow} presents the obtained results of DDNet and DDNet augmented with flow information. Some classes like sky, sign and pedestrians greatly benefit from the flow information, with sign and sky improving more than six percentage points, which is a strong gain for the already high quality segmentation obtained by DD-Net with RGB.

\begin{table*}[h]
\centering
\caption[Comparison between DDNet and DDNet augmented with flow]{Comparison between DDNet and DDNet augmented with flow on reduced version of the Cityscapes dataset. As seen the flow augmented version of DD-Net present superior results in the majority of the classes and an improvement in the mean IoU metric close to $1.5$ percentage points.}
\label{cityscapes_flow}
\begin{tabular}{l||c|c|c|c|c|c|c|c|c|c||c}
Method & \rot{Sky} &  \rot{Building} &  \rot{Road} &  \rot{Sidewalk} &  \rot{Cyclist} &  \rot{Vegetation} &  \rot{Pole} &  
\rot{car} &  \rot{Sign} &  \rot{Pedestrian} & \rot{mIoU} \\
\toprule
\toprule
DD-Net   & $85.27$ & $88.99$ & $\mathbf{97.22}$ & $\mathbf{80.06}$ & $\mathbf{67.64}$ & $90.32$ & $\mathbf{52.50}$ & $91.08$ & $60.94$ & $68.95$ & $78.30$\\

DD-Net+FLow   & $\mathbf{91.60}$ & $\mathbf{89.90}$ & $97.15$ & $79.63$ & $66.03$ & $\mathbf{90.70}$ & $52.13$ & $\mathbf{92.32}$ & $\mathbf{68.10}$ & $\mathbf{70.10}$ & $\mathbf{79.77}$\\

\bottomrule
\end{tabular}
\end{table*}

\begin{figure*}[h!]
   \centering
    \begin{subfigure}{.245\linewidth}
       \includegraphics[width=\linewidth]{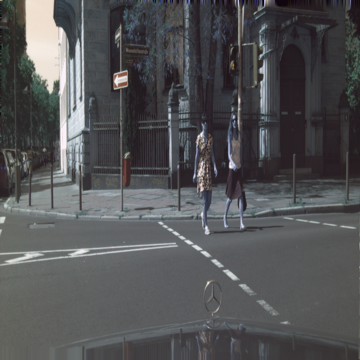}
    \end{subfigure}     
    \begin{subfigure}{.245\linewidth}    
       \includegraphics[width=\linewidth]{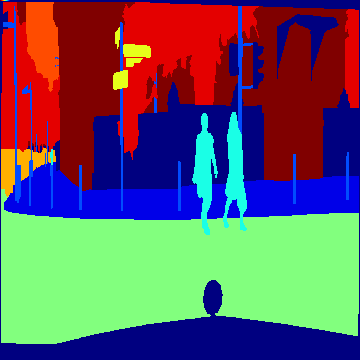}
    \end{subfigure}     
    \begin{subfigure}{.245\linewidth}
       \includegraphics[width=\linewidth]{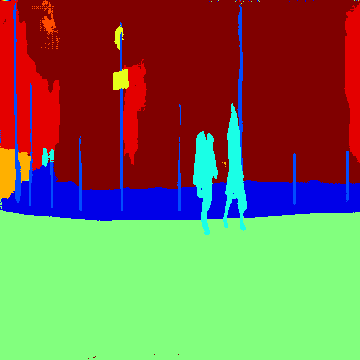}
    \end{subfigure}     
    \begin{subfigure}{.245\linewidth}    
       \includegraphics[width=\linewidth]{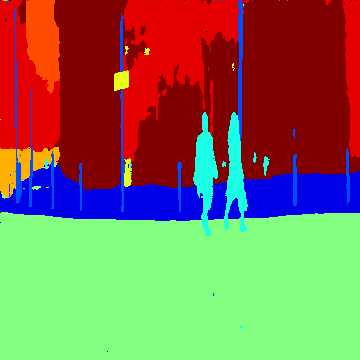}
    \end{subfigure}     \\
    \begin{subfigure}{.245\linewidth}
       \includegraphics[width=\linewidth]{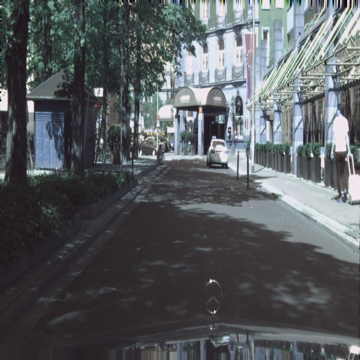}
    \end{subfigure}     
    \begin{subfigure}{.245\linewidth}    
       \includegraphics[width=\linewidth]{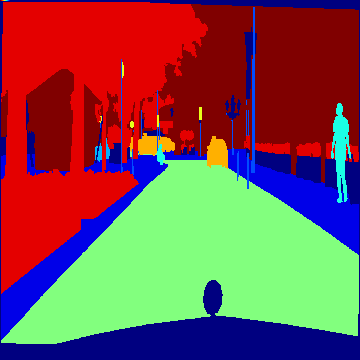}
    \end{subfigure}     
    \begin{subfigure}{.245\linewidth}
       \includegraphics[width=\linewidth]{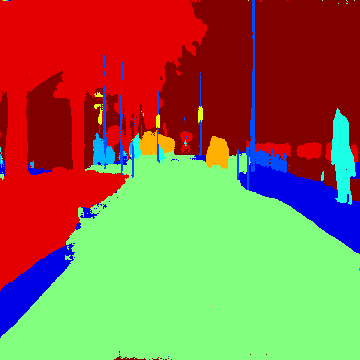}
    \end{subfigure}     
    \begin{subfigure}{.245\linewidth}    
       \includegraphics[width=\linewidth]{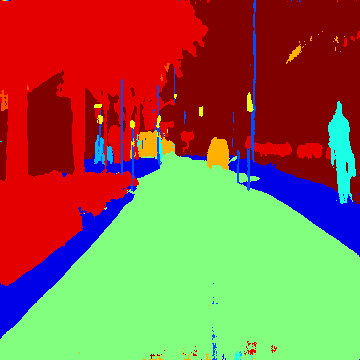}
    \end{subfigure}     \\     
    \begin{subfigure}{.245\linewidth}
       \includegraphics[width=\linewidth]{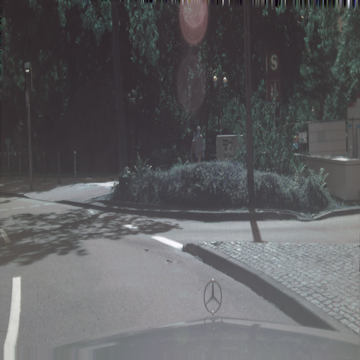}
    \end{subfigure}     
    \begin{subfigure}{.245\linewidth}    
       \includegraphics[width=\linewidth]{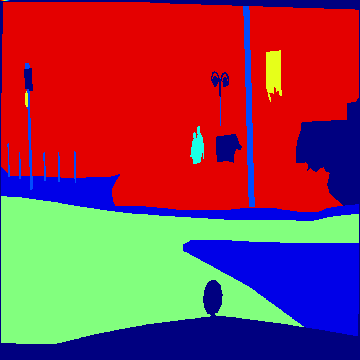}
    \end{subfigure}     
    \begin{subfigure}{.245\linewidth}
       \includegraphics[width=\linewidth]{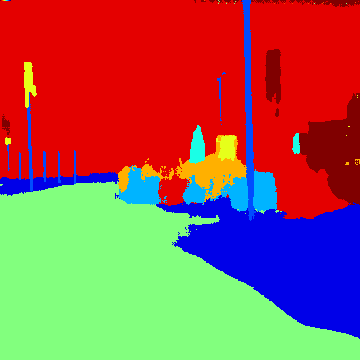}
    \end{subfigure}     
    \begin{subfigure}{.245\linewidth}    
       \includegraphics[width=\linewidth]{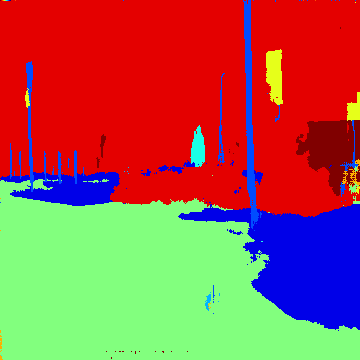}
    \end{subfigure}     \\ 
    \begin{subfigure}{.245\linewidth}
       \includegraphics[width=\linewidth]{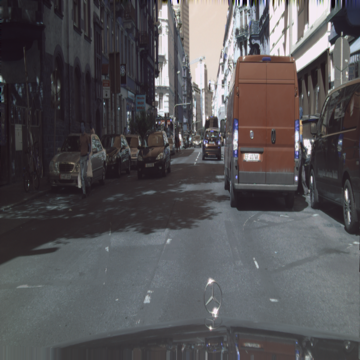}
    \end{subfigure}     
    \begin{subfigure}{.245\linewidth}    
       \includegraphics[width=\linewidth]{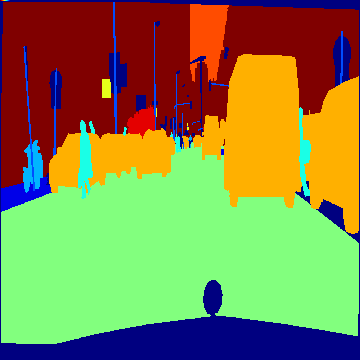}
    \end{subfigure}     
    \begin{subfigure}{.245\linewidth}
       \includegraphics[width=\linewidth]{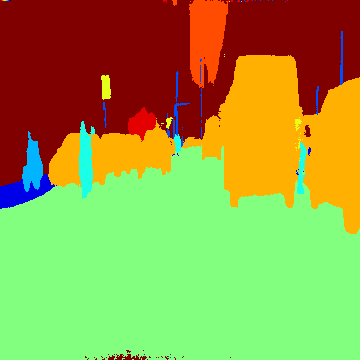}
    \end{subfigure}     
    \begin{subfigure}{.245\linewidth}    
       \includegraphics[width=\linewidth]{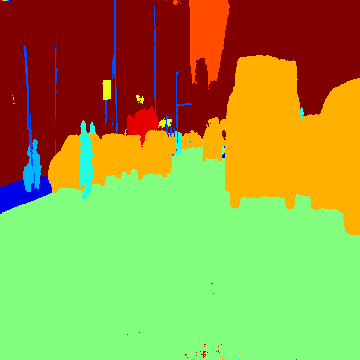}
    \end{subfigure}     \\ 
	\caption[DD-Net and Flow augmented DD-Net]{Segmentation results between DD-Net and Flow augmented DD-Net. The first column shows input images, the second column the groundtruth, the third one the prediction from DD-Net without augmentation and the fourth column the DD-Net with flow augmentation. The first row is an example where the pedestrian class is better segmented by the flow augmented DD-Net. In the second row it can be seen that DD-Net alone presents less false positive detections for the building class but perform worst in the sidewalk segmentation. The third row is an inverse instance where DD-Net presents a considerable area in the center of the image with false positive detections of multiple classes, while the flow augmented version correctly segmentation that region as vegetation. The last row presents an example where the flow augmented version of our approach is capable of better segment the poles, which is an extremely challenging class.}
	\label{fig:flow_no_flow_DDNet}
\end{figure*}

A qualitative comparison between our approach and its motion augmented version is presented in Figure \ref{fig:flow_no_flow_DDNet}. The columns are organized as follows, input image, groundtruth, DD-Net prediction and flow augmented DD-Net respectively. The first row is an example where the inclusion of motion can provide better segmentation for classes in which motion can be better capture, that is the case for the pedestrian class. The second row shows that even with an inferior mean IoU for the sidewalk class, in the flow augmented version, the specific instance presents superior segmentation values for this class when compared to the standard RGB DD-Net. The third row exemplifies the lower false positive detection rate from the augmented version of the network, the vegetation class suffers from inconsistent labelling in the center of the image for the RGB version of our technique. The last row presents an example where the flow augmented version of our approach is capable of better segment the poles, which is an extremely challenging class. The main outcome from the optical flow augmentation to our approach is that motion related classes, like pedestrians, can benefit from that. However for the specific case of the cityscapes dataset the gain is limited by the similar motion pattern of the forward moving car. We believe that datasets with richer motion patterns will further benefit from this strategy.

\section{Conclusions}
\label{sec:conclusion}

In this paper, we have proposed a new decoder that is composed by a set of shallow small networks for semantic segmentation. The new decoder consists of a new topology of skip connections, namely backward and stacked residual and with a novel weight function, in which we aim to re-balance classes to increase the attention of the networks to under represented objects. The ablation study shows that the design options effectively capture more information, are less conditioned to false positive detection and can produce a more efficient architecture for the given depth. Additionally, we show that a compact version of our approach is capable of iterative frame rate with minimum reduction of segmentation capabilities. Our experimental results show that our approach yields state-of-the-art results on the most relevant benchmarks for robotics and that motion clues can be used as extra input values to further improve segmentation.   

%\begin{table*}[ht]
%\centering
%\begin{tabular}{l||c|c|c|c|c|c|c|c|c|c||c}
%Method & \rot{Sky} &  \rot{Building} &  \rot{Road} &  %\rot{Sidewalk} &  \rot{Cyclist} &  \rot{Vegetation} &  %\rot{Pole} &  
%\rot{car} &  \rot{Sign} &  \rot{Pedestrian} & \rot{mIoU} \\
%\toprule
%\toprule
%AdapNet  \parencite{valada17icra}  & $87.13$ & $83.14$ & $94.45$ & $68.93$ & $52.36$ & $85.72$ & $39.44$ & $84.16$ & $50.73$ & $47.81$ &  $69.39$\\
%DD-Net   & $85.27$ & $88.99$ & $\bf{97.22}$ & $\bf{80.06}$ & $\bf{67.64}$ & $90.32$ & $\bf{52.50}$ & $91.08$ & $60.94$ & $68.95$ & $78.30$\\

%DD-Net + Auxiliary Flow  & $\bf{91.60}$ & $\bf{89.90}$ & $97.17$ & $79.63$ & $66.03$ & $\bf{90.70}$ & $52.13$ & $\bf{92.32}$ & $\bf{68.32}$ & $\bf{70.10}$ & $\bf{79.77}$\\

%\bottomrule
%\end{tabular}
%\end{table*}

%\begin{table*}[ht]
%\centering
%\begin{tabular}{l||c|c||c}
%Method & Static &  Dynamic & mIoU \\
%\toprule
%\toprule
%SmsNet \parencite{Vertens17iros}  & $84.27$ & $75.31$ & $79.79$\\
%DD-Net + Auxiliary Flow  & $\bf{92.36}$ & $75.16$ & $\bf{83.76}$\\

%\bottomrule
%\end{tabular}
%\end{table*}

\begin{acknowledgements}
This work was partially funded by the Freiburg Graduate School of Robotics and a research project with Valeo Vision Systems, Ireland. 
\end{acknowledgements}

% % BibTeX users please use one of
% \bibliographystyle{spbasic}      % basic style, author-year citations
% %\bibliographystyle{spmpsci}      % mathematics and physical sciences
% %\bibliographystyle{spphys}       % APS-like style for physics
% \bibliography{references}   % name your BibTeX data base

\printbibliography

% Non-BibTeX users please use
%\begin{thebibliography}{}
%
% and use \bibitem to create references. Consult the Instructions
% for authors for reference list style.
%
%\bibitem{VGG}
% Format for Journal Reference
%Karen Simonyan and Andrew Zisserman, Very Deep Convolutional Networks for Large-Scale Image Recognition, ICLR, 2015
% Format for books
%\bibitem{RefB}
%Author, Book title, page numbers. Publisher, place (year)
% etc
%\end{thebibliography}

\end{document}